\documentclass[journal]{IEEEtran}
\ifCLASSINFOpdf
\else
\fi

\usepackage{microtype}
\usepackage{graphicx}
\usepackage{subfigure}
\usepackage{booktabs} 
\usepackage{amsmath,amssymb,amsfonts}
\usepackage{bm}

\usepackage{color}
\usepackage{algorithm}
\usepackage{algorithmic}
\usepackage{array}
\usepackage{multirow}

\usepackage{makecell}

\usepackage[table]{xcolor}

\newtheorem{theorem}{Theorem}

\newtheorem{lemma}{Lemma}
\newtheorem{corollary}{Corollary}
\newtheorem{definition}{Definition}

\newtheorem{remark}{Remark}

\begin{document}
\title{Leveraged Matrix Completion with Noise}

\author{
	    Xinjian Huang,
        Weiwei Liu,~\IEEEmembership{Senior Member,~IEEE,}
        Bo Du,~\IEEEmembership{Senior Member,~IEEE,}\\
        Dacheng Tao,~\IEEEmembership{Fellow,~IEEE}
\IEEEcompsocitemizethanks{
	    \IEEEcompsocthanksitem Xinjian Huang is with the School of Cyber Science and Engineering, Nanjing University of Science and Technology, Nanjing, China, and also with the School of Computer Science, Wuhan University, Wuhan, China.\protect\\
	    \indent Weiwei Liu and Bo Du are with the School of Computer Science, National Engineering Research Center for Multimedia Software, Institute of Artificial Intelligence and Hubei Key Laboratory of Multimedia and Network Communication Engineering, Wuhan University, Wuhan, China.\protect\\
		\indent Dacheng Tao is with the	School of Computer Science,	 in the Faculty of Engineering, at the University of Sydney, 6 Cleveland St, Darlington, NSW 2008, Australia. \protect\\
       \indent Co-corresponding Authors:
        liuweiwei863@gmail.com (Weiwei Liu); \protect\\
        dubo@whu.edu.cn (Bo Du).
}
}

\markboth{Journal of \LaTeX\ Class Files,~Vol.~*, No.~*, MONTH~YEAR}%
{Shell \MakeLowercase{\textit{et al.}}: Bare Demo of IEEEtran.cls for Computer Society Journals}

\IEEEtitleabstractindextext{%
\begin{abstract}
Completing low-rank matrices from subsampled measurements has received much attention in the past decade. Existing works indicate that $\mathcal{O}(nr\log^2(n))$ datums are required to theoretically secure the completion of an $n \times n$ noisy matrix of rank $r$ with high probability, under some quite restrictive assumptions: (1) the underlying matrix must be incoherent; (2) observations follow the uniform distribution. The restrictiveness is partially due to ignoring the roles of the leverage score and the oracle information of each element. 
In this paper, we employ the leverage scores to characterize the importance of each element and significantly relax assumptions to: (1) not any other structure assumptions are imposed on the underlying low-rank matrix; (2) elements being observed are appropriately dependent on their importance via the leverage score. 
Under these assumptions, instead of uniform sampling, we devise an ununiform/biased sampling procedure that can reveal the ``importance'' of each observed element. Our proofs are supported by a novel approach that phrases sufficient optimality conditions based on the Golfing Scheme, which would be of independent interest to the wider areas. Theoretical findings show that we can provably recover an unknown $n\times n$ matrix of rank $r$ from just about $\mathcal{O}(nr\log^2 (n))$ entries, even when the observed entries are corrupted with a small amount of noisy information. The empirical results align precisely with  our theories.

\end{abstract}

}

\maketitle

\IEEEdisplaynontitleabstractindextext

%
\IEEEpeerreviewmaketitle

\section{Introduction}\label{sec:Introduction}


\IEEEPARstart{M}{atrix} completion \cite{Candes2009717,Nie202212042,Li2022a,Zhang20192580,He2022,Foucart20211264,Li2022,Li2023,Tsakiris2023}
is a fundamental task aimed at recovering a low-rank matrix from a small subset of its elements. This problem has attracted considerable attention from researchers due to its wide range of applications across various fields, including image processing \cite{Bouwmans20181427,Lu20191859}, recommendation systems \cite{Kang2016185}, multi-task learning \cite{Mao2020,Mao2020a}, dimensionality reduction \cite{Lu20151900}, clustering and localization in sensor networks \cite{Wang2017981}, drug-target interactions prediction \cite{Li2021}, and traffic speed estimation \cite{Wang20232023}, to name a few.
However, in real-world application, noise is an inevitable factor during the data acquisition process, as it reflects the uncertainty of the environment and/or the measurement processes.
For example, in the context of the Netflix problem, users' ratings are uncertain \cite{Sharma20193223}. Similarly, in the positioning problem, local instances are imperfect \cite{Liao20131511}. Additionally, in the acquisition of the functional MRI, its signal may be contaminated by subject motion artifacts \cite{Balachandrasekaran2022}.

To formulate the above-mentioned problem precisely, imagine that we consider the scenario where our objective is to recover an unknown low-rank matrix $\hat L\in\mathbb R^{n_1\times n_2}$ from a collection of partially observed and corrupted entries $X$ as follows:
\begin{align}
	X_{ij} = \hat L_{ij} + \hat S_{ij}, \quad (i,j)\in O,
\end{align}
where $\hat S\in\mathbb R^{n_1\times n_2}$ is a matrix representing random noises, and we only observe entries over an index  subset $O\subseteq[n_1]\times[n_2]$ with $[n_1]:=\{1,\ldots,n_1\}$ and $[n_2]:=\{1,\ldots,n_2\}$. The aim is to reliably recover $\hat L$ given the incomplete and even grossly corrupted data.

Cand{\`e}s et al. \cite{Candes2010925} first show that this goal can be achieved by means of a principled convex program
\begin{align}\label{equ:targetopt}
	\begin{split}
		\min_{L,S}\quad  &\|L\|_*+\lambda\|S\|_1\\
		\text{s.t.}\quad &\mathcal{P}_O(X) = \mathcal{P}_O(L)+S,
	\end{split}
\end{align}
where $\|L\|_*$ denotes the nuclear norm (the sum of the singular values) of $L$, $\|S\|_1= \sum_{i,j}|S_{ij}|$ denotes the entrywise $l_1$ norm and $\mathcal{P}_O(\cdot)$ is the orthogonal projection operator which keeps the elements in $O$ invariant and $0$ otherwise; $X\in\mathbb{R}^{n_1\times n_2}$ is the matrix containing the known entries (with values known only on $O$).
Specifically, they suggest that at least $\mathcal{O}(nr\log^2(n))$ observations are required to theoretically secure completing an $n \times n$ noisy matrix of rank $r$ with high probability under the assumptions that the observed elements follow the \emph{uniform} distribution and that the low-rank matrix to be recovered should satisfy \emph{incoherence} property \cite{Chen20152909}. 
The incoherence property, while useful in various applications such as nonnegative matrix factorization \cite{Lu20185248} and face recognition \cite{Wei20143294}, necessitates that the row or column spaces of the considered data should be diffuse \cite{Candes2010925,Chen2011873}.

Unfortunately, these restrictive assumptions do	not hold in the majority of real-world applications. For example, in the sensor network localization problem \cite{Oh2010}, information observed by the key nodes is evidently more informative than that observed by the edge nodes \cite{Shi2016}.
Thus, it is necessary to consider the ``importance'' of elements being observed during the sampling processing, which may result in non-uniform sample processing. In this context, leverage score can offer a promising solution. It was first introduced to detect outliers in regression diagnostics \cite{Hoaglin197817}, and has proven successful in analyzing the large-scale data and implementing randomized matrix due to its capacity to measure the correlation of the dominant subspace with the canonical basis \cite{Mahoney2010123}.
For an element in the matrix, its ``importance'' thus can be characterized by the sum of the leverage scores of its corresponding row and column.

In this paper, we address two interconnected questions simultaneously: \emph{(1) can we devise a specific sampling strategy suitably dependent on the ``importance'' of each observed entry?} and \emph{(2) does this sampling strategy works effectively even under noisy cases?}
Indeed, we show that achieving reasonably accurate matrix completion from \emph{noisy sampled entries} is feasible, given that the sampling process incorporates a biased distribution based on the \emph{leverage scores} of the target low-rank matrix.
Specifically, we consider a noisy low-rank matrix completion problem, in which each sampled element follows a specific distribution defined by its leverage scores, and the incoherence property of the low-rank matrix is nonessential in our approach. Under these reasonable conditions, we devise a more compact
sample complexity $\mathcal{O}(nr\log^2(n))$ with respect to noisy low-rank matrix completion. This sampling upper bound is derived mainly on the basis of Golfing Scheme (an elegant technique to construct dual certificates \cite{Gross20111548,Recht2011,Candes20111}) and several concentration inequalities involving two norms defined by the leverage score. Our theoretical findings are further validated through empirical results.

The main contributions of this paper are summarized as follows:
\begin{itemize}
	\item Our theoretical results show that if the sampling probability follows a biased distribution determined by the row and column leverage scores of the underlying matrix, only $\mathcal{O}(nr\log^2(n))$ observed entries are needed to exactly recover the underlying low-rank matrix with high probability, even in the presence of corruption in a subset of the observed entries (Theorem \ref{alg:procedure}).
	\item Our proof techniques introduce a new method for formulating sufficient optimality conditions (Lemma \ref{lem:duallem}) based on the Golfing Scheme by constructing a particular matrix-valued random process that converges to the dual certificate $Y$ (Section \ref{sec:dualconstruct}).
	\item Two norms, which involve leverage scores, are presented to better characterize the probability of each element being observed. Furthermore, several concentration properties \cite{Tropp20151} and upper bounds with respect to these norms are also rigorously derived (Section \ref{sec:dualcert}).
\end{itemize}

The rest of the paper is organized as follows. In Section \ref{sec:relatedwork}, we briefly review some related work. Section \ref{sec:preliminary} elaborates on some preliminaries about the problem we considered. Our main theoretical findings are presented in Section \ref{sec:mainresults}. Proof of our theoretical results are provided in Section \ref{sec:4proofmain}. Section \ref{sec:5exp} reports our empirical results. Section \ref{sec:6conclusion}  concludes the work. The proofs of some theorems are available in the Appendix.

\section{Related Work} \label{sec:relatedwork}
Matrix completion has been extensively studied in both theory and algorithm due to its wide application in numerous scenarios. If $S$ in the convex optimization (\ref{equ:targetopt}) equals to $0$, then it involves the following realistic matrix completion problem:
\begin{equation}\label{equ:MCnuclear}
	\begin{split}
		\min_L\quad &\|L\|_*\\
		\text{s.t.}\quad &\mathcal P_O(L) = \mathcal P_O(X),
	\end{split}
\end{equation}

Next, we first give some notations utilized in this paper, and then briefly review some literature on matrix completion with and without noises from both theory and algorithm perspectives.\vspace{2mm}

\textbf{Notations:} $X_{ij}$ denotes the $(i,j)$-th element of a matrix $X\in\mathbb{R}^{n_1\times n_2}$. $X_{i\cdot}$ and $X_{\cdot j}$ are the $i$-th row and $j$-th column of $X$, respectively. $X^*$ denotes the transpose of $X$. There are five norms associated with a matrix $X$: $\|X\|_F$ denotes the Frobenius norm, $\|X\|_*$ denotes the nuclear norm,  $\|X\|$ denotes the spectral norm and $\|X\|_1$ and $\|X\|_\infty$ represent the $l_1$ and $l_\infty$ norms of the long vector stacked by $X$. The inner product between two matrices is $\langle X,Y\rangle=\text{trace}(X^*Y)$. We denote by $\text{Range}(\mathcal{P})$ the range of an operator $\mathcal{P}$. A linear operator $\mathcal{A}$ acts on the space of matrices and $\|\mathcal{A}\|$ denotes the operator norm given by $\|\mathcal{A}\|=\sup_{\{\|X\|_F=1\}}\|\mathcal{A}(X)\|_F$.

\subsection{Theory}
In general, restoration of a matrix from a small amount of observations is impossible. However, if the unknown matrix has low-rank structure, then accurate and even exact recovery is possible.

Cand{\`e}s and Recht \cite{Candes2009717} provide the first algorithm and theoretical guarantees for low-rank matrix completion, where they show that the nuclear norm minimization problem (\ref{equ:MCnuclear}) works when the low-rank matrix is incoherent and the sampling process is uniform and independent of the matrix, obtaining that the underlying low-rank matrix can be exactly recovered with high probability from only $\mathcal{O}(n^{1.25}r\log(n))$ entries. Subsequent works have refined provable completion results that $\mathcal{O}(nr\log^2(n))$ entries are needed for incoherent matrices recovery under the uniform random sampling model \cite{Gross20111548,Candes2010,Recht2011}. The same complexity has also  been obtained under a more suitable setting. Chen et al. \cite{Chen20152999} prove that any coherent matrix can be exactly recovered from $\mathcal{O}(nr\log^2(n))$ entries if the random sampling processing follows a specific biased distribution. If the row space is coherent and the column space is still incoherent, Krishnamurthy and Singh \cite{Krishnamurthy2013} establish that merely $\mathcal{O}(nr^{1.5}\log(r))$ entries can exactly recover the low-rank matrix with high probability under an adaptive sampling strategy. Further, they also prove that $\mathcal{O}(nr^{1.5}\text{ploy}\log(n))$ entries can recover the noisy low-rank matrix. For life-long or online matrix completion, Balcan and Zhang \cite{Balcan20162955} establish an optimal guarantee that exactly recovers an $\mu$-incoherent matrix by probability at least $1-\delta$ with sample complexity $\mathcal{O}(\mu rn\log(\frac{r}{\delta}))$ in the context of sparse random noise. More recently, based on leave-one-out technique, Ding and Chen \cite{Ding20207274} show that only $\mathcal{O}(\mu r \log(\mu r) n \log(n))$ observations are suffice to recover an $n\times n$ $\mu$-incoherence matrix of rank $r$ by using nuclear norm minimization methods. Although \cite{Ding20207274} and \cite{Balcan20162955} establish a tighter sampling bound for noise-free and noisy matrix completion, respectively, these results are developed under the $\mu$-incoherence assumption, a stronger assumption than coherence considered in this paper.

To make a clear comparison, Table \ref{tab:com} summarizes these theoretical results. Our theoretical findings establish that the required number of observations for successfully recovering a noisy low-rank matrix is $\mathcal{O}(nr\log^2(n))$, which is in the same order as the state-of-the-art results when no corruptions exist and more reasonable assumptions are satisfied.
\begin{table}[!htb]
    \centering
	\renewcommand\arraystretch{1.3}
	\caption{Short summary of several upper bounds for low-rank matrix completion under different settings.}
	\label{tab:com}
	\begin{tabular}{c c c c c}
		\hline
		\hline
		Ref.         &Noisy       &Sampling           &Assumption &Upper Bound \\
		\hline
        \cite{Candes2009717} &no &uniform &incoherence &$\mathcal{O}(n^{1.2}r\log(n))$\\
	    \cite{Candes2010} &no &uniform &incoherence &$\mathcal{O}(nr\text{ploy}\log(n))$\\
	    \cite{Candes2010925} &yes &uniform &incoherence &$\mathcal{O}(nr\log^2(n))$\\
	    \cite{Balcan20162955} & yes & uniform &incoherence &$\mathcal{O}(\mu rn\log(\frac{r}{\delta}))$\\
	    \cite{Ding20207274} &no &uniform &incoherence &$\mathcal{O}(\mu r \log(\mu r) n \log(n))$\\
    	\cite{Chen20152999} &no &non-uniform &coherence &$\mathcal{O}(nr\log^2(n))$\\
		\cite{Krishnamurthy2013} & no & adaptive & coherence &$\mathcal{O}(nr^{1.5}\log(r))$\\
		\cite{Krishnamurthy2013} & yes & adaptive & coherence &$\mathcal{O}(nr^{1.5}\log(n))$\\
		ours &yes &non-uniform &coherence &$\mathcal{O}(nr\log^2(n))$\\
		\hline
	\end{tabular}
\end{table}

\subsection{Algorithm}
Existing matrix completion methods can be generally classified into three categories: regularization-based methods, matrix factorization-based methods and others.

\noindent \textit{$\left.1 \right)$ Regularization-based method}

The regularization-based method for matrix completion involves using singular values or their variations to construct different norms to surrogate rank function. The most famous surrogation for the rank function is the nuclear norm, which corresponds to the nuclear norm-based matrix completion problem (\ref{equ:MCnuclear}). The singular value threshold (SVT) method \cite{Cai20101956} and the inexact augmented Lagrange multiplier (IALM) method  \cite{Lin2013a,Lin2011612} are two representative methods designed to solve the optimization problem (\ref{equ:MCnuclear}). Afterwards, some other regularizes are also developed to surrogate the low-rank function, such as weighted nuclear norm \cite{Gu2017183,Gu20142862}, truncated nuclear norm \cite{Hu20132117,Oh2016744}, Shatten $p$-norm \cite{Nie2012655,Fan2019}, weighted Shatten $p$-norm \cite{Xie20164842} and Schatten capped $p$-norm \cite{Li2022b}. Specific algorithms are rigorously devised concerning these different norms. Note that these norms can be used not only in matrix completion, but also in other computer vision tasks, such as image denoising \cite{Huang2020637,Huang2020271,Hu20191487}, image impainting\cite{Li20195962}, etc.

The key challenge of the regularization-based method is the computationally expensive calculation of singular values, especially in large scale cases. To improve the computational efficiency, matrix approximation is often used as an efficient alternative, thus inducing the following subsection.

\noindent \textit{$\left.2 \right)$ Matrix factorization-based method}

This kind of method is to factorize or approximate the original matrix by the product of two or more small-sized matrices. Compared with the regularization-based methods, this type of method is usually computationally cheap and less memory-consuming.

The maximum margin factorization (MMF) method \cite{Srebro20041329} is a representative factorization-based method. However, empirical studies show that the MMF method only can obtain sub-optimal solutions. The low-rank matrix fitting (LMaFit) method \cite{Wen2012333} is devised by constructing a nonlinear successive over-relaxation algorithm. LMaFit only requires solving a series of linear least-squares problems, which enables this method to handle the large-scale problem well. Recently, Shang et al. \cite{Shang201553, Shang20182066} propose a bilinear factorization method to solve the low-rank matrix recovery problem. The method approximates the original matrix by two small-scaled matrix and some specific norms, i.e., nuclear norm, are imposed on each small-scaled matrix to induce the low-rank property. Factor group-sparse regularization is also developed for completing matrices \cite{Fan2019}. Additionally, it proves that the Schatten-$p$ norm is the sum of two group-sparse norms, which greatly enhances computational efficiency \cite{Fan2019}. Based on the factorization framework, Wang et al. \cite{Wang20231521} propose a novel robust matrix completion scheme via using the truncated-quadratic loss function, and half-quadratic theory is adopted for its optimization.

\noindent \textit{$\left.3 \right)$ Others}

Besides the aforementioned methods, some other methods are also extensively developed. Based on a sum of multiple orthonormal side information and nuclear-norm regularization, Ledent et al. \cite{Ledent20232259} propose an interpretable approach to matrix completion with a provable  convergence. To handle the data matrices with non-linear structures, Fan et al. propose non-linear matrix completion (NLMC), which extends the conventional matrix completion method to non-linear structures \cite{Fan2018378}. For high-rank matrix completion problem, a novel online method is proposed by using kernel trick, where it maps the data into a high dimensional polynomial feature space \cite{Fan20198690}. The proposed online method enjoys much lower space and time complexity since the data admit a low dimensional subspace in this feature space. Based on the correntropy criterion, He et al. \cite{He2020181} proposed a half-quadratic alternating steepest descent (HQ-ASD) algorithm for a robust matrix completion problem (\ref{equ:targetopt}). To further utilize the smooth Riemannian manifold of a  matrix with a fixed-rank, Riemannian optimization \cite{Hu2020199} was introduced to accelerate the optimization process \cite{Vandereycken20131214}. Based on linear latent variable models, deep matrix factorization (DMF) \cite{Fan201834} is proposed for nonlinear matrix completion. Recently, learning-based methods have also been introduced to the task of completion \cite{Wu2022}. For distributed matrix completion problem, \cite{Abubaker2023} proposes a framework for scaling stratified SGD through significantly reducing the communication overhead. Zhang et al. employ the alternating direction method of multiplier (ADMM) with two dual variables to optimize the generalized nonconvex nonsmooth low-rank matrix recovery problems \cite{Zhang2022}.

\section{Preliminary} \label{sec:preliminary}
Suppose matrix $X\in\mathbb{R}^{n_1\times n_2}$ is the sum of an underlying low-rank matrix $\hat{L}\in\mathbb{R}^{n_1\times n_2}$ and a sparse ``noises'' matrix $\hat{S}\in\mathbb{R}^{n_1\times n_2}$. We consider the following problem: suppose we only observe a subset $O\subseteq[n_1]\times[n_2]$ of the entries of $X$ ; the remaining entries are unobserved. Our goal is to exactly and provably recover $\hat{L}$ from partially observed entries with noise. Formally, we focus on the following noisy matrix completion problem:
\begin{align}\label{equ:targetopt0}
	\begin{split}
		(\hat{L},\hat{S})=\text{arg}\min_{L,S}\quad  &\|L\|_*+\lambda\|S\|_1\\
		\text{s.t.}\quad &\mathcal{P}_O(X) = \mathcal{P}_O(L)+S,
	\end{split}
\end{align}
where $S$ is supported on the index matrix $\Omega\subset O$, $\mathcal{P}_O(X)$ is the matrix obtained by setting the entries of $X$ that are outside the observed set $O$ to zero and $\lambda$ is a parameter that trades off between these two elements of the objective function. The value of $\lambda$ is chosen for a theoretical guarantee of exact recovery in Theorem \ref{thm:mainthm}. The nuclear norm is used as a convex surrogate for the rank of a matrix and the $l_1$ norm is used as a convex surrogate for its sparsity \cite{Recht2010471}.


Notice that the observed data is $\mathcal{P}_O(L)+S$, where $O\subseteq[n_1]\times[n_2]$ and $S$ is supported on $\Omega\subset O$. We assume that the index set $O$ of the observation data is obtained by non-uniform sampling with probability $p_{ij}$ \footnote{Note that the value of $p_{ij}$ is determined by the leverage scores of each datum. Details can be seen in Section IV to follow.}. Random uniform corruption of these observations with probability $q$ yields the index set $\Omega$ of the ``noise'' matrix $S$. Specifically, the index matrix $O,\Omega$ and sparse ``noise'' matrix $S$ satisfy the following model:

\noindent\textbf{Model 1}\label{mod:1}
\begin{itemize}
	\item $L$ is supported by $O\subseteq[n_1]\times[n_2]$; $O$ is determined by Bernoulli sampling with \emph{non-uniformly} probability with $p_{ij}$, denoted as $O\sim \text{Ber}(p_{ij})$. That is to say, $p_{ij}$ represents the probability that the $(i,j)$-th entry of $L$ to be observed or sampled.
	\item Assume that $\Omega$ is \emph{uniformly} sampled from $O$ with probability $q$ and that ``noise'' matrix $S$ is supported by $\Omega$. In other words, Given $(i,j)\in O$, we have $\mathbb P((i,j)\in\Omega|(i,j)\in O)=q$. This implies that $\Omega$ is determined by Bernoulli sampling with non-uniformly probability $qp_{ij}$, denoted as $\Omega\sim \text{Ber}(qp_{ij})$.
	\item Define $\Gamma := O/\Omega$. We then have $\Gamma\sim \text{Ber}(p_{ij}(1-q))$.
	\item Define $\text{sgn}(S)=\mathcal P_\Omega(K)$, where $K\in\mathbb R^{n_1\times n_2}$ and its entries are either $1$ or $-1$.
\end{itemize}

We assume $L$ is of rank $r$ and its reduced singular value decomposition (SVD) is denoted as $L=U\Sigma V^*$, where $U\in\mathbb{R}^{n_1\times r}$, $\Sigma\in\mathbb{R}^{r\times r}$ and $V\in\mathbb{R}^{n_2\times r}$. We next provide the definition of \emph{leverage scores}, which is used to determine the non-uniformly sampling process.

\begin{definition}
	(Leverage Scores) For a real-valued matrix $L\in\mathbb{R}^{n_1 \times n_2}$ with rank $r$, its SVD  is $U\Sigma V^*$. Then its row leverage score $\mu_i(L)$ for any row $i$ and column leverage score $\nu_j(L)$ for any column $j$ are defined as
	\begin{align*}
		&\mu_i = \frac{n_1}{r}\|UU^*e_i\|^2, \quad \text{for}~ i = 1,2,\ldots,n_1,\\
		&\nu_j = \frac{n_2}{r}\|VV^*e_j\|^2, \quad \text{for}~ j = 1,2,\ldots,n_2,
	\end{align*}
	where $e_i$ is the $i$-th canonical basis vector in Euclidean space (the vector with all entries equal to $0$ but the $i$-th equal to $1$) with appropriate dimension.
\end{definition}

Note that the leverage scores of the matrix $X$ are non-negative, and are functions of the column and row spaces of $L$. The standard coherence parameter $\mu$ of $L$ used in the previous literature \cite{Candes2009717,Chen20152909} corresponds to a global upper bound on the leverage scores, i.e., $\mu=\max_{i,j}\{\mu_i(L),\nu_j(L)\}$. Clearly, standard coherence parameters characterize the quality of a matrix from a holistic perspective, while leverage scores from a local perspective. Therefore, the leverage scores can be considered as localized versions of the standard coherence parameter.

Two norms ($\mu{(\infty)}$-norm and $\mu{(\infty,2)}$-norm) with respect to leverage scores are needed in the following concentration properties establishment. The $\mu{(\infty,2)}$-norm of a matrix $Z\in\mathbb{R}^{m\times n}$ is defined as
\begin{equation*}
	\|Z\|_{\mu(\infty,2)}:=\max_{a,b}\{\sqrt{\frac{m}{\mu_ar}}\|Z_{a\cdot}\|_2, \sqrt{\frac{n}{\nu_br}}\|Z_{\cdot b}\|_2\},
\end{equation*}
which is the maximum of the weighted column and row norms of $Z$. $\mu{(\infty)}$-norm of $Z$ is defined as
\begin{equation*}
	\|Z\|_{\mu{(\infty)}}:=\max_{a,b}|Z_{ab}|\sqrt{\frac{m}{\mu_a r}}\sqrt{\frac{n}{\nu_b r}},
\end{equation*}
which is the weighted element-wise magnitude of $Z$. $\|Z\|_{\mu(\infty,2)}$ and $\|Z\|_{\mu{(\infty)}}$ are exactly norms. Detailed proofs can be seen in the Appendix \ref{app:normproof}.

\section{Leveraged Matrix Completion with Noise}\label{sec:mainresults}
\subsection{Main Results}
To facilitate the derivation of the upper bound for noisy matrix completion by using leverage scores, Model 1 is transformed into the following model in an equivalent manner.

\noindent\textbf{Model 2}\label{mod:2}
\begin{itemize}
	\item Fix an $n\times n$ matrix $K$, whose entries are either $1$ or $-1$.
	\item Define two independent random subset of $[n_1]\times[n_2]$: $\Gamma'\sim\text{Ber}(p_{ij}(1-2q))$ and $\Omega'\sim\text{Ber}(\frac{2p_{ij}q}{1-p_{ij}+2qp_{ij}})$. Let $O:=\Gamma'\cup\Omega'$, it is easy to verify that $O\sim\text{Ber}(p_{ij})$.
	\item Define an random matrix $W\in\mathbb R^{n_1\times n_2}$ with independent entries $W_{ij}$ satisfying $\mathbb P(W_{ij}=1)=\mathbb P(W_{ij}=-1)=\frac{1}{2}$.
	\item Define $\Omega''\subset\Omega'$, where $\Omega'':=\{(i,j)|W_{ij}=K_{ij}, (i,j)\in \Omega'\}$. Then define $\Omega:=\Omega''/\Gamma'=\frac{1}{2}\Omega'/\Gamma'$ and $\Gamma:=O/\Omega$.
	\item Let $\text{sgn}(S) :=\mathcal{P}_\Omega(K)$.
\end{itemize}
Clearly, in both Model 1 and Model 2, if we fix $(O,\Omega)$, the whole setting is deterministic. Therefore, the probability of $(\hat{L},\hat{S})=(L,S)$ is determined by the joint distribution of $(O,\Omega)$. Besides, it is easy to verify that the joint distribution of $(O,\Omega)$ in the two models is identical.

We are now ready to state our main results presented as follows. For simplicity, results provided here are on the basis of square matrix with size $n\times n$; similar results can be extended to a general rectangle case in the same fashion.
\begin{theorem}\label{thm:mainthm}
	Under the Model 2, if each element $(i,j)$ is independently observed with probability $p_{ij}$ and satisfies
	\begin{align*}
		p_{ij}&\geq \max\left\{c_p\frac{(\mu_i+\nu_j)r\log^2(n)}{n},\frac{1}{n^5}\right\},
	\end{align*}
	$q\leq c_q$ and $\lambda=\frac{1}{24\sqrt{n\log n}}$, then $(\hat{L},\hat{S})$ is the unique optimal solution to the problem (\ref{equ:targetopt}) with probability at least $1-Cn^{-5}$ for a positive constant $C$, provided that the positive constants $c_p$ is sufficiently large and $c_q$ is sufficiently small.
\end{theorem}

\emph{Proof sketch:} Due to space constraints, we present only the outline of the theorem's proof here. The full proof is outlined in Section \ref{sec:4proofmain} and the Appendix. The high-level roadmap of the proof follows a standard approach: to demonstrate that $\hat{L}$ represents the unique optimal solution to problem \eqref{equ:targetopt}, it is necessary to construct a dual certificate $Y$ that adheres to specific sub-gradient optimality conditions. Specifically, by optimization theory, we first establish the first order subgradient sufficient conditions for Problem (\ref{equ:targetopt}) in Lemma \ref{lem:duallem}. Subsequently, employing standard duality theory and the Golfing Scheme, we derive a dual certificate $Y$ that satisfies conditions \eqref{equ:cond1}-\eqref{equ:cond4} with high probability under certain conditions. Finally, we validity that $Y$ satisfies all conditions \eqref{equ:cond1}-\eqref{equ:cond4} under specific assumptions. Differ from the previous work that bound the $\ell_\infty$ norm $\|Z\|_\infty:=\max_{i,j}|Z_{i,j}|$ of a random matrix $Z$, we instead bound two weighted norms, $\mu{(\infty)}$-norm and $\mu{(\infty,2)}$-norm, to derive several inequalities via Bernstein inequality.

\begin{remark}
	In Theorem \ref{thm:mainthm},
	$c_p$ and $c_q$ are universal positive constants, whose values can be well-designed during the proofs of the corresponding inequalities or lemmas. Details can be see at Section V and the Appendices.
\end{remark}

\begin{remark}\label{remark:1}
    The power of our results is that one can recover a low-rank matrix with rank $r$ from nearly minimal number of samples in the order of $\mathcal{O}(nr\log^2 n)$ even when a constant proportion of these samples has been corrupted. From experimental results, we know that this corruption proportion empirically approximates $0.6$ by using leverage sampling, while $0.18$ by using uniform sampling (see Section \ref{sec:5exp} for detail). Moreover, this theorem implies that elements with higher leverage scores should be sampled with higher probability. Informally, elements with higher leverage scores have more ``important information'' of the matrix, thereby tolerating larger noise density.
\end{remark}

\begin{remark}
	Theorem 1 states us that only $\mathcal O(nr\log^2(n))$ datums can theoretically guarantee the exact recovery of a noisy low-rank matrix. This upper bound on sampling is consistent with to the findings in \cite{Chen20152999}. 
	However, our results are based on the assumption of mild noise contamination in the sampled data, while the results in \cite{Chen20152999} are based on the assumption of no noise contamination. Consequently, our results can be viewed as an extension of the work in \cite{Chen20152999}, with broader applicability.
\end{remark}

%

\subsection{Leverage Sampling in Practice}
So far, we have established that one can exactly recover an arbitrary $n\times n$ rank-$r$ matrix from just about $\mathcal{O}(nr\log^2(n))$ observations if sampled in accordance with the leverage scores. However, partial observations (sometimes even corrupted) result in the lack of a prior knowledge about the leverage scores, preventing the implementation of leveraged sampling in practice.
	
To overcome the restriction on applications, we find that $X$ and $L$, intuitively, share the same row and column space and that perturbation or sparse noise $S$ does not change the leverage scores if $\text{range}(X)=\text{range}(L)$. These findings provide a possible way to calculate leverage scores via the observed matrix $X$. Formally, the following Theorem can theoretically ensure that leverage scores of $L$ can be approximated via observed matrix $X$.

\begin{theorem}\label{thm:leverageclose}
		Let $X$, $L$ and $S$ are defined in (\ref{equ:targetopt}), with $\text{Range}(L)=\text{Range}(X)$ and $\delta=\|S\|\|L^\dag\|\leq \frac{1}{2}$, where $L^\dag$ denotes the Moore-Penrose inverse of $L$. Then
		\begin{align}
			\frac{\|\tilde{\mu}_i-\mu_i\|}{\mu_i}\leq (2\sqrt{\frac{1-\mu_i}{\mu_i}}+\frac{\delta}{\mu_i})\delta,\quad\text{for}~ i=1,\ldots,m, \label{equ:nearmu}
		\end{align}
		and
		\begin{align}
			\frac{\|\tilde{\nu}_j-\nu_j\|}{\nu_j}\leq (2\sqrt{\frac{1-\nu_j}{\nu_j}}+\frac{\delta}{\nu_j})\delta,\quad\text{for}~ j=1,\ldots,n, \label{equ:nearnu}
		\end{align}
		where $\tilde{\mu}_i$ and $\tilde{\nu}_j$ denote the $i$-th row and $j$-th column leverage score of $X$; ${\mu}_i$ and ${\nu}_j$ denote the $i$-th row and $j$-th column leverage score of $L$.
\end{theorem}
	\emph{Proof sketch:} Based on principal angle theory \cite{Holodnak20151143}, it can be observed that the contribution of noise $S$ to the range of $L$ is limited. Then by the oracle information of the leverage scores and Theorem 2.4 in \cite{Holodnak20151143}, our results hold under appropriate assumptions.
		Detailed proofs can be seen in the Appendix \ref{app:a}.

\begin{remark}
	From Theorem \ref{thm:leverageclose}, we get that the relative leverage score difference between $L$ and $X$ is bounded by a very small constant under some suitable conditions. Thus we can directly use the observation data to calculate the leverage score by existing methods \cite{Chen20152999,Eftekhari2018581}. For the number of samples, we can rank the leverage scores in a descending order and then select the top $N$ samples, as long as $N$ satisfies the sampling upper bound provided in Remark \ref{remark:1}.
\end{remark}

Based on Theorem \ref{thm:leverageclose} and the existing methods \cite{Chen20152999,Eftekhari2018581}, we propose Algorithm \ref{alg:procedure} for leveraged matrix completion with noise: first estimating the leverage scores of the noisy matrix from a small number of uniform samples, then using these estimated leverage scores to select the remaining samples. Specifically, given a total budget of $N$ samples, draws a subset $O$ uniformly without replacement such that $|O|=\theta N$, where $\theta\in[0,1]$ denotes the fraction of the budget to estimate the leverage scores of the underlying matrix.
Then take a rank-r approximation to $\mathcal{P}_O(X)$, $\tilde{U}\tilde{\Sigma}\tilde{V}^T$, obtaining the estimated leverage scores  $\tilde{\mu}_i=\mu_i(\tilde{U}\tilde{\Sigma}\tilde{V}^T)$ and $\tilde{\nu}_j=\nu_j(\tilde{U}\tilde{\Sigma}\tilde{V}^T)$. Last, generate the remaining $(1-\theta)N$ samples by sampling without replacement with distribution $p_{ij}\propto {(\tilde\mu_i+\tilde\nu_j)r\log^2(n)}/{n}$, obtaining the new set of samples $\tilde{O}$. Take a union of $O$ and $\tilde{O}$ to construct the observation data $\mathcal{P}_{O\cup\tilde{O}}(X)$ as constraints in robust matrix completion problem (\ref{equ:targetopt}).

\begin{algorithm}
	\caption{Leverage Sampling for Matrix Completion with Noise}
	\label{alg:procedure}
	\begin{algorithmic}[1]
		\REQUIRE {Noisy data $X$, rank $r$, sampling budget $N$, $\theta$}
		\STATE {Draw a subset $O$ by sampling uniformly without replacement such that $|O|$=$\theta N$}.
		\STATE {Compute rank-$r$ approximation to $\mathcal{P}_O(X)$, $\tilde{U}\tilde{\Sigma}\tilde{V}^T$}.
		\STATE {Calculate the estimated leverage scores  $\tilde{\mu}_i=\mu_i(\tilde{U}\tilde{\Sigma}\tilde{V}^T)$ and $\tilde{\nu}_j=\nu_j(\tilde{U}\tilde{\Sigma}\tilde{V}^T)$}.
		\STATE {Generate the remaining $(1-\theta)N$ samples by sampling without replacement with distribution $p_{ij}\propto {(\tilde\mu_i+\tilde\nu_j)r\log^2(n)}/{n}$}.
		\STATE {Replace $\mathcal{P}_O$ with $\mathcal{P}_{O\cup\tilde{O}}(X)$ in problem (\ref{equ:targetopt}) to obtain
		\begin{equation}\label{equ:probest}
			\begin{split}
			\min_{L,S}&\quad\|L\|_*+\lambda\|S\|_1,\\
			\text{s.t.}&\quad \mathcal{P}_{O\cup\tilde{O}}(X) = \mathcal{P}_{O\cup\tilde{O}}(L)+S.
			\end{split}
		\end{equation}
            	}
        \STATE{Solve problem (\ref{equ:probest}) to obtain $\hat{L}$}.
		\ENSURE {Recovered matrix $\hat{L}$}.
	\end{algorithmic}
\end{algorithm}

\section{Proof of Theorem \ref{thm:mainthm}}\label{sec:4proofmain}
Our analysis of non-uniform error bound is based on leverage scores, where $\mu{(\infty)}$-norm and $\mu{(\infty,2)}$-norm are utilized to establish concentration properties and bounds. Our proof includes two main steps: (1) deriving the sufficient condition for the optimality of Problem (\ref{equ:targetopt}) and (2) constructing such a dual certificate by Golfing Scheme \cite{Gross20111548}.

We first introduce a few additional notations. We define a subspace $T$ that share either the same column space or the row space as ${L}$: $T = \{ UX^*+YV^*:X\in\mathbb R^{n_2 \times r}, Y \in\mathbb R^{n_1\times r}\}$. As a matter of fact, $T$ is the tangent space with respect to $\mathcal P(\text{rank}(M))$ at $M$, where $\mathcal P(k):= \{M\in\mathbb R^{n_1\times n_2}|\text{rank}(M)\leq k\}$ \cite{Chandrasekaran2011572}. $T$ induces a projection $\mathcal{P}_T$ given by $\mathcal{P}_T(M)=UU^*M+MVV^*-UU^*MVV^*$. $T^\bot$ denotes the complement subspace to $T$, also induces a projection $\mathcal{P}_{T^\bot}$ with $\mathcal{P}_{T^\bot}=(I-UU^*)M(I-VV^*)$. $\mathcal{P}_\Omega(M)$ is the matrix with $(\mathcal{P}_\Omega(M))_{ij}=M_{ij}$ if $(i,j)\in\Omega$ and zero otherwise. $\mathcal{P}_{\Omega^{c}}(M):=M - \mathcal{P}_\Omega(M)$.
Instead of denoting several positive constant $C_0,C_1,c,\ldots$, we just use $C,C^{'}$, whose values may change from line to line.
We will use the phrase ``with high probability" to mean with high probability at least $1-Cn^{-5}$.

\subsection{Sufficient Condition for Optimality}
We first derive the first order subgradient sufficient conditions for Problem (\ref{equ:targetopt}) as below:

\begin{lemma} \label{lem:duallem}
	If $p_{ij}\geq \max\{c_p\frac{(\mu_i+\nu_j)r\log^2(n)}{n},\frac{1}{n^5}\}$,  $\lambda=\frac{1}{24\sqrt{n\log n}}$ and there exists a dual variable $Y\in\mathbb{R}^{n\times n}$ satisfying
	\begin{align}\label{equ:dualcondition}
		\left\{
		\begin{array}{lll}
			\|\mathcal P_{T}(Y+\lambda\mathcal P_{\Omega^{'}}(W)-UV^*)\|_F\leq\frac{\lambda}{n^3} \\
			\|\mathcal P_{T^\bot}(Y+\lambda\mathcal P_{\Omega^{'}}(W))\|\leq\frac{1}{4} \\
			\|\mathcal{P}_{\Gamma^{'}}(Y)\|_\infty\leq\frac{\lambda}{4} \\
			\mathcal{P}_{\Gamma^{'^c}}(Y)=0,
		\end{array}
		\right.
	\end{align}
	then the solution $(\hat{L},\hat{S})$ is the unique optimal solution to the original optimization problem (\ref{equ:targetopt}).
\end{lemma}
\emph{Proof.} The proof details of this lemma can be found in the Appendix \ref{app:b}.

\subsection{Construction of the Dual Certificate} \label{sec:dualconstruct}
Since $O:=\Gamma^{'}\cup\Omega^{'}$, we know that $\mathcal{P}_{O/\Gamma^{'}}(W)=\mathcal{P}_{\Omega^{'}/\Gamma^{'}}(W)$. From Model 2, we know that the distribution of $(\Gamma^{'},\mathcal{P}_{\Omega^{'}}(W))$ and $(\Gamma^{'},2\mathcal{P}_{\Omega^{'}/\Gamma^{'}}(W)-\mathcal{P}_\Omega^{'}(W))$ are same.

Suppose there exist $Y_1$ and $Y_2$ satisfying
\begin{align*}
	&\|\mathcal{P}_T(Y_1+Y_2)+\mathcal{P}_T[\lambda(2\mathcal{P}_{\Omega^{'}/\Gamma^{'}}(W)-2UV^*)]\|_F\\
	&\!\!\leq\|\mathcal{P}_T(Y_1)+\mathcal{P}_T[\lambda(2\mathcal{P}_{\Omega^{'}/\Gamma^{'}}(W)-\mathcal{P}_{\Omega^{'}}(W)-UV^*)]\|_F\!\!\\
	&+\|\mathcal{P}_T(Y_2)+\mathcal{P}_T(\lambda\mathcal{P}_{\Omega^{'}}(W)-UV^*)\|_F\\
	&\leq\frac{\lambda}{n^3}+\frac{\lambda}{n^3}= \frac{2\lambda}{n^3},
\end{align*}
and
\begin{align*}
	&\|\mathcal{P}_{T^\bot}(Y_1+Y_2)+\mathcal{P}_{T^\bot}[\lambda(2\mathcal{P}_{\Omega^{'}/\Gamma^{'}}(W))]\|\\
	&\leq\|\mathcal{P}_{T^\bot}(Y_1)+\mathcal{P}_{T^\bot}[\lambda(2\mathcal{P}_{\Omega^{'}/\Gamma^{'}}(W)-\mathcal{P}_\Omega^{'}(W))]\|,\\
	&+\leq\|\mathcal{P}_{T^\bot}(Y_2)+\mathcal{P}_{T^\bot}(\lambda\mathcal{P}_{\Omega^{'}}(W))\|\\
	&\leq \frac{1}{4}+\frac{1}{4}= \frac{1}{2}.
\end{align*}
Then we know that $Y=\frac{Y_1+Y_2}{2}$ satisfies the condition (\ref{equ:dualcondition}). To prove Theorem \ref{thm:mainthm}, we need to prove that there exists $Y$ satisfying
\begin{align}
	&\|\mathcal{P}_T(Y)+\mathcal{P}_T(\lambda\mathcal{P}_{\Omega^{'}}(W)-UV^*)\|_F\leq \frac{\lambda}{n^3}, \label{equ:cond1}\\
	&\|\mathcal{P}_{T^\bot}(Y)+\mathcal{P}_{T^\bot}(\lambda\mathcal{P}_{\Omega^{'}}(W))\|\leq\frac{1}{4},\label{equ:cond2}\\
	&\|\mathcal{P}_{\Gamma^{'}}(Y)\|_\infty\leq\frac{\lambda}{4}, \label{equ:cond3}\\
	&\mathcal{P}_{\Gamma^{'^c}}(Y)=0,\label{equ:cond4}
\end{align}
with high probability under the assumptions of Lemma \ref{lem:duallem}.

Notice that $\Gamma^{'}\sim\text{Ber}(p_{ij}(1-2q))$. Suppose that $\rho$ satisfies $1-p_{ij}(1-2q)=(1-\frac{p_{ij}(1-2q)}{6})^2(1-\rho)^{t-2}$, where $t=\lfloor5\log n+1\rfloor$. We know that $\rho\leq \frac{C\rho}{\log n}$. Define $\rho_1=\rho_2=\frac{p_{ij}(1-2q)}{6}$, $\rho_3=\ldots=\rho_t=\rho$. Let $\Gamma^{'}=\Gamma_1\cup\Gamma_2\cup\ldots\cup\Gamma_t$, where $\Gamma_k\sim\text{Ber}(\rho_k)$ independently.

Construct
\begin{align}\label{equ:constructy}
	\left\{
	\begin{array}{lll}
		X_0=\mathcal{P}_T(UV^*-\lambda\mathcal{P}_{\Omega^{'}}(W))\\
		X_k=(\mathcal{P}_T- \frac{1}{\rho_k}\mathcal{P}_T  \mathcal P_{\Gamma_{k}}  \mathcal{P}_T)X_{k-1}\\
		Y=\sum_{k=1}^{t}\frac{1}{\rho_k}\mathcal P_{\Gamma_{k}}X_{k-1}.
	\end{array}
	\right.
\end{align}

By this construction, we see that
\begin{equation}\label{equ:e2}
X_k=\mathcal{P}_T(X_k),\quad k=0,1,\ldots,t,
\end{equation}
which implies that $X_k$ is in the range of $\mathcal{P}_T(X_k)$.

\subsection{Validity of the Dual Certificate}\label{sec:dualcert}
We next to show that $Y$ satisfies all the constraints (\ref{equ:cond1})-(\ref{equ:cond4}) simultaneously under our assumptions. The inequality (\ref{equ:cond4}) is immediately held by the construction of $Y$. Before validating the constraints (\ref{equ:cond1})-(\ref{equ:cond3}), we present some Lemmas.

\begin{lemma}(Matrix Bernstein Inequality \cite{Tropp2012389})\label{lem:bernstein}
	Let $X_1,X_2,\ldots,X_n\in\mathbb{R}^{n_1\times n_2}$ be independent zero mean random matrices. Suppose
	\begin{equation*}
		\max\{\|\mathbb E\sum_{k=1}^{n}X_kX_k^*\|,\|\mathbb E\sum_{k=1}^{n}X_k^*X_k\|\}\leq\sigma^2
	\end{equation*}
	and $\|X_k\|\leq B$ almost surely for all $k$. Then we have
	\begin{equation*}
		\mathbb{P}\{\|\sum_{k=1}^{n}X_k\|\geq t\}\leq(n_1+n_2)\exp(\frac{-t^2/2}{Bt/3+\sigma^2}).
	\end{equation*}
	As a consequence, for any $c>0$, we have
	\begin{equation*}
		\|\sum_{k=1}^{n}X_k\|\leq 2\sqrt{c\sigma^2\log(n_1+n_2)}+cB\log(n_1+n_2),
	\end{equation*}
	with probability at least $1-(n_1+n_2)^{1-c}$.
\end{lemma}

\begin{lemma} \cite{Chen20152999}\label{lem:lem1}
	If $p_{ij}\geq c_p\frac{(\mu_i+\nu_j)r\log^2(n)}{n}$ for all $(i,j)$ and $q\leq c_q$, then with high probability
	\begin{equation}
	\|\frac{1}{(1-2q)p_{ij}}\mathcal{P}_T\mathcal P_{\Gamma}\mathcal{P}_T-\mathcal{P}_T\|\leq\frac{1}{2},
	\end{equation}
	provided that $c_p$ is sufficiently large and $c_q$ is sufficiently small.
\end{lemma}

From Lemma \ref{lem:lem1}, we have
\begin{equation*}
	\|X_k\|_F\leq\frac{1}{2}\|X_{k-1}\|_F,\quad k=1,2,\ldots,t,
\end{equation*}
with high probability, provided $c_p$ is sufficiently large and $c_q$ is sufficiently small. Therefore, it easy to obtain
\begin{equation}\label{equ:i5}
\|X_k\|_F\leq\frac{1}{2^k}\|X_{0}\|_F,\quad k=1,2,\ldots,t.
\end{equation}

\begin{lemma}\label{lem:lem3}
	Suppose $Z$ is a fixed $n\times n$ matrix and $Z\in\text{Range}(\mathcal{P}_T)$. If $p_{ij}\geq c_p\frac{(\mu_i+\nu_j)r\log^2(n)}{n}$ for all $(i,j)$ and $q\leq c_q$, then with high probability
	\begin{align*}
		&\|(\frac{1}{(1-2q)p_{ij}}\mathcal{P}_T\mathcal P_{\Gamma}\mathcal{P}_T-\mathcal{P}_T)Z\|_{\mu(\infty,2)}\\
		&\leq\frac{1}{2}(\|Z\|_{\mu(\infty)}+\|Z\|_{\mu(\infty,2)}),
	\end{align*}
	provided that $c_p$ is sufficiently large and $c_q$ is sufficiently small.
\end{lemma}
\emph{Proof.} The proof details of this lemma is provided in the Appendix \ref{app:c}.

\begin{lemma}\label{lem:lem4}
	Suppose $Z$ is a fixed $n\times n$ matrix and $Z\in\text{Range}(\mathcal{P}_T)$. If $p_{ij}\geq c_p\frac{(\mu_i+\nu_j)r\log^2(n)}{n}$ for all $(i,j)$ and $q\leq c_q$, then with high probability
	\begin{align*}
		&\|(\frac{1}{(1-2q)p_{ij}}\mathcal{P}_T\mathcal P_{\Gamma}\mathcal{P}_T-\mathcal{P}_T)Z\|_{\mu(\infty)}\leq\frac{1}{2}\|Z\|_{\mu(\infty)},
	\end{align*}
	provided that $c_p$ is sufficiently large and $c_q$ is sufficiently small.
\end{lemma}
\emph{Proof.} The proof details of this lemma is provided in the Appendix \ref{app:d}.

\begin{corollary}\label{cor:cor1}
	Suppose $Z$ is a fixed $n\times n$ matrix and $Z\in\text{Range}(\mathcal{P}_T)$. If there exists a $\alpha>0$ such that $p_{ij}\geq c_0\sqrt{\frac{(\mu_i+\nu_j)r}{n\alpha^2}}\geq c\frac{(\mu_i+\nu_j)r\log^2(n)}{n}$ for all $(i,j)$, provided $c_0$ and $c$ is sufficiently large and $q$ is sufficiently small. Then with high probability
	\begin{align*}
		&\|(\frac{1}{(1-2q)p_{ij}}\mathcal{P}_T\mathcal P_{\Gamma}\mathcal{P}_T-\mathcal{P}_T)Z\|_{\mu(\infty)}\leq\frac{\alpha}{2}\|Z\|_{\mu(\infty)}.
	\end{align*}
\end{corollary}

Note that this Corollary can be seen as a generalization of Lemma 3.1 in \cite{Candes20111}.

Armed with the above lemmas, dual certifications of (\ref{equ:cond1})-(\ref{equ:cond3}) can be extensively conducted, which is provided in the Appendix \ref{app:e}.

\section{Experiments}\label{sec:5exp}
In this section, we provide numerical experiments for solving extensive synthetic and real-world problems to demonstrate the effectiveness of our proposed leveraged sampling strategy and the correctness of the theoretical findings.

\subsection{Synthetic Experiment}
We first conduct an experiment by considering a simulated task on artificially generated data, whose goal is to restore a clean matrix from partially observed entries with noises. We declare that a trial is successful if $\|\hat{L}-L\|_F/\|L\|_F\leq 0.05$. Robust bilinear factorization (RBF) \cite{Shang201553}, a classical decomposition-based method for robust matrix completion problems, is utilized to solve problem (\ref{equ:probest}).

The low-rank matrix $L$ is constructed by $L=X_1X_2^*$, where the entries of $X_1,X_2\in\mathbb{R}^{n\times r}$ are independently sampled from Gaussian distribution $\mathcal{N}(0,\frac{1}{n^2})$. In order to verify that sampling by leverage scores can bear more corruptions, we study the following two types of models.
\begin{itemize}
	\item Uniform sampling + Uniform corruption (UU): each entry is sampled with equal probability $p_{ij}=p$, for all $(i,j)\in [n]\times[n]$; each observed entries are corrupted by $S$, where $\mathbb{P}(S_{ij}=1)=\mathbb{P}(S_{ij}=-1)=\frac{1}{2}$ and $\mathbb{E}(S_{ij})=q$, for all $(i,j)\in O$.
	\item Leveraged sampling + Uniform corruption (LU): each entry is sampled with probability $$p_{ij}=p\frac{n^2\sqrt{\mu_i+\nu_j}}{\sum_{i,j}\sqrt{\mu_i+\nu_j}}$$for all $(i,j)\in [n]\times[n]$; $S$ is the same as UU.
\end{itemize}

\begin{figure}[t]
	\centering
	\includegraphics[scale=0.45]{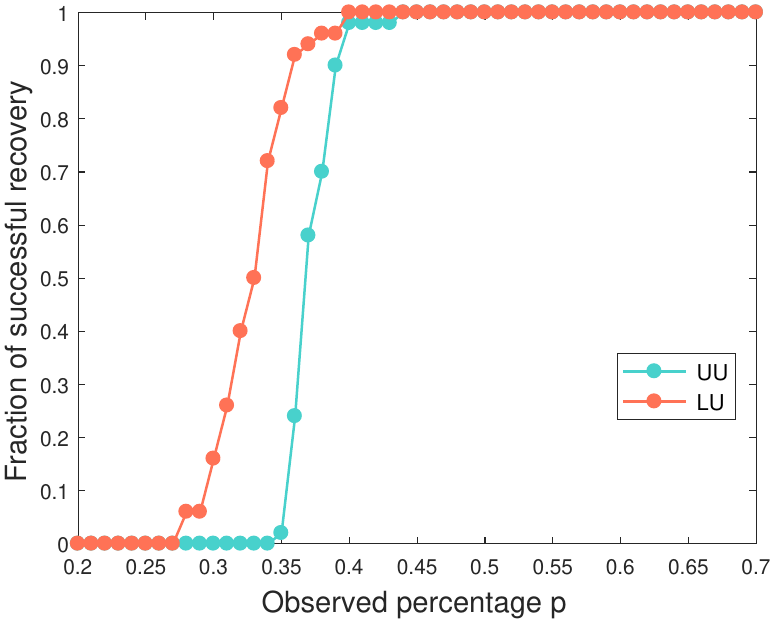}
	\caption{Ratio of successful recovery versus observed percentage with $q=0.1$.}
	\label{fig:obspercent}
\end{figure}

In LU model, the probability is adaptive to the leverage scores.

We first demonstrate that exact recovery of a low rank matrix with noise not only depends on the percentage of the entries, but also how entries are observed. For UU and LU models, we set $n=1000$, $r=10$. Figure \ref{fig:obspercent} shows the successful frequency versus the observed percentage $p$, where we fix $q=0.1$. For each value of $p$, we perform 50 trials of independent observations and error corruptions and count the number of successes. We observe that the LU model outperforms the UU model. With same ratio of successes, LU needs less observed entries; with same observed entries, LU can obtain higher ratio of success. This is because LU is based on leverage scores, which can be used to characterize the importance of each element in a matrix.

\begin{figure}[t]
	\centering
	\includegraphics[scale=0.45]{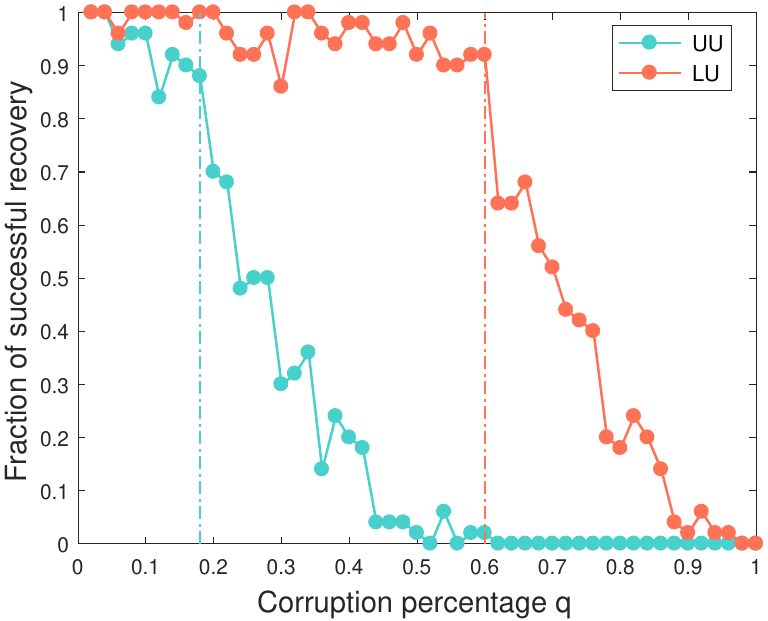}
	\caption{Ratio of successful recovery versus corruption percentage with $p=0.4$.}
	\label{fig:corruption}
\end{figure}

We next study the the influence of the corruption percentage on the ratio of successes. We also set $n=1000$, $r=10$. Figure \ref{fig:corruption} shows the successful frequency versus the corruption percentage $q$, where we fix $p=0.4$. For each value of $q$, the ratio of successes are obtained in the same way as above. We observe that the LU model performs significantly higher robustness to noise. Specifically, in this setting, the capacity of noise immunity increases about $2.33$ times, from $0.18$ to $0.6$. This is mainly because sampling by leverage scores can reveal ``dominating'' elements of a low-rank matrix.

\subsection{Collaborative filtering}
In this subsection, we propose to verify the effectiveness of our proposed leverage sampling method on real-world applications -- collaborative filtering; this is a technique for some recommender systems, aiming to recommend movies to its users. Formally, it predicts the unknown preference of a user on a set of unrated items according to other similar users or items.

\emph{MovieLens}\footnote{https://grouplens.org/datasets/movielens/} and \emph{Jester}\footnote{https://goldberg.berkeley.edu/jester-data/} are two widely used datasets for recommender systems \cite{Harper20152160,Srebro2003720}. For MovieLens, we select \emph{MovieLens 100K} (ML-100K) and \emph{MovieLens 1M} (ML-1M) in our experiments. ML-100K contains $100,000$ ratings for $1,682$ movies by $943$ users and the ratings range from $1$ to $5$. ML-1M contains $1,000,209$ ratings for about $3,900$ movies by $6,040$ users and the ratings range from $1$ to $5$.
\emph{Jester} dataset is a joke rating dataset. It consists of three rating matrices, namely Jester-1, Jester-2 and Jester-3. Ratings of these datasets are continuous real values ranging from $-10$ to $10$. Dimension descriptions for each dataset are provided in Table \ref{tab:dataset_info}. Details of these datasets can be seen on their official website.

\begin{table}
	\arrayrulecolor{black}
	\centering
	\color{black}
	\caption{Information of the real datasets utilized in our experiments.}
	\label{tab:dataset_info}
	\begin{tabular}{lllll}
		\hline
		Datasets & \#Users & \#Items & \#Rated Items & Range      \\
		\hline
		ML-100K  & 943    & 1,682   & 100,000      & {[}1,5{]}  \\
		ML-1M    & 6,040   & 3,706  & 1,000,209    & {[}1,5{]}  \\
		Jester-1 & 24,983  & 100    & 100,000      & [-10,10] \\
		Jester-2 & 23,500  & 100    & 100,000     & [-10,10] \\
		Jester-3 & 24,938  & 100    & 60,000      & [-10,10]\\
		\hline
	\end{tabular}
\end{table}

For each dataset, $X_{clean}$ denotes the original clean data. For \emph{MovieLens}, similar to \cite{Alquier20192117}, we add artificial noises by randomly changing $5\%$ of ratings that are equal to $5$ to $1$, and randomly changing $5\%$ of ratings that are equal to $1$ to $5$, thereby constructing the noisy data $X_{noise}$.
For \emph{Jester} datasets, we first remove the first column of each data matrix. And then $10\%$ of the ratings are randomly selected for the rounding operation, thereby obtaining the noisy data $X_{noise}$.
Training data $X_{train}$ for each dataset is constructed in two different ways: \emph{leverage sampling} by Algorithm \ref{alg:procedure} and \emph{uniform sampling}. Testing data is constructed by $X_{test}:=\{x_{ij}|x_{ij}\in X_{clean}~ \text{and}~ x_{ij}\notin X_{train}\}$.

The ratio of training set to testing set is defined as $\varrho_1 := {|X_{train}|}:{|X_{test}|}$. Also, we define the ratio of the sampling budget $N$ to the size of the training set as $\varrho_2={N}/{|X_{train}|}$. Root Mean Squared Error (RMSE) is utilized to measure the accuracy of the recovered results, which is defined on the test set:
\begin{equation*}
	RMSE = \sqrt{\frac{\sum_{(i,j)\in TestSet}(X_{ij}-L_{ij})^2}{|TestSet|}},
\end{equation*}
where $TestSet$ denotes the index set of the testing set $X_{test}$ and $|TestSet|$ is the total number of ratings in $TestSet$.

Two decomposition-based robust matrix completion algorithms, RBF \cite{Shang201553} and HQASD \cite{He2020181}, are utilized to solve problem (\ref{equ:probest}) and we set $r=7$ in these two methods.

For \emph{MovieLens} dataset, we set $\varrho_1=9:1$ and $8:2$. The average RMSE results on ML-100K and ML-1M are reported over 10 independent trials and are shown in the Table \ref{tab:exp100k} and Table \ref{tab:expmo1m}, respectively. For \emph{Jester} dataset, we provide the results with parameter $\varrho_1=9:1$ over $10$ trials in Table \ref{tab:expjester}.  In these tables, ``uni'' means uniform sampling and ``lev'' means leverage sampling. We can see that leverage sampling outperforms random uniform sampling both on RBF and HQASD methods. We also see that parameter $\theta=0.7$ leads to the best performance, which is consistent with the empirical findings in \cite{Chen20152999}.

Furthermore, empirical results also demonstrate the effectiveness and the superiority of our proposed algorithm, implying that just a small fraction of observations, even corrupted, to rate a few selected movies according to the estimated leverage scores obtained by previous samples have the potential to greatly improve the quality of the recovered preference matrix.

\begin{table}
\arrayrulecolor{black}
\centering
\caption{Average RMSE of $10$ trials by different sampling strategies on ML-100K dataset under different parameters setting.}
\label{tab:exp100k}
\begin{tabular}{ l|l|l|lll l}
	\hline
	$\varrho_1$  &$\varrho_2$       &$\theta$        &RBF-uni           &RBF-lev   &HQASD-uni  &HQASD-lev\\
	
	\hline
	\multirow{8}{*}{9:1} &
	\multirow{4}{*}{0.9} &0.6 &0.8294   &0.8109& 0.8100 &0.8073\\
	                         &  &0.7   &0.8229& 0.8033 &0.8020 & 0.7981\\
	                        &   &0.8   &0.8297& 0.8174 &0.8044 & 0.8006\\
	                           &  &0.9 & 0.8324 &0.8263 & 0.8221 &0.8147\\
	                           \cline{2-7}
   &\multirow{4}{*}{0.8} &0.6   &0.9097 & 0.9001 &0.9037 & 0.8941\\
   						&  &0.7   &0.9032 & 0.8974 &0.8999 & 0.8957\\
					   &   &0.8   &0.9103 &0.9047 &0.9026 &0.9003 \\
					   &  & 0.9 & 0.9174 & 0.9096 & 0.9117 & 0.9097\\
   						\cline{2-7}
	\hline\hline
	\multirow{8}{*}{8:2} &
	\multirow{4}{*}{0.9} &0.6   &0.8522& 0.8501 &0.8529 & 0.8484\\
						&  &0.7   &0.8526 & 0.8473 &0.8511 & 0.8466\\
						&   &0.8   &0.8530 & 0.8512 &0.8500 & 0.8493\\
						&  & 0.9 &0.8687& 0.8660 &0.8624& 0.8617\\
						\cline{2-7}
						&\multirow{4}{*}{0.8}
						&0.6   &0.9605& 0.9574 &0.9473& 0.9531\\
						&  &0.7   &0.9563 & 0.9521 &0.9462& 0.9499\\
						&   &0.8   &0.9620 & 0.9607 &0.9420 &0.9587\\

						&  & 0.9 &0.9907 & 0.9883 &0.9638& 0.9701\\
						\cline{2-7}
	\hline
\end{tabular}
\end{table}

\begin{table}
	\centering
	\caption{Average RMSE of $10$ trials by different sampling strategies on ML-1M dataset under different parameters setting.}
	\label{tab:expmo1m}
	\begin{tabular}{ l|l|l|lll l}
		\hline
		$\varrho_1$  &$\varrho_2$       &$\theta$        &RBF-uni           &RBF-lev   &HQASD-uni  &HQASD-lev\\
		
		\hline
		\multirow{8}{*}{9:1} &
		\multirow{4}{*}{0.9} &0.6   &0.9583&0.9422  &0.9466 & 0.9407\\
		&  &0.7   & 0.9527 & 0.9401 &0.9432 &0.9388\\
		&   &0.8   &0.9602 &0.9516  &0.9556 & 0.9473\\
		&  & 0.9 & 0.9731 & 0.9662 &0.9703 & 0.9534\\
		\cline{2-7}
		&\multirow{4}{*}{0.8} &0.6   &1.247& 1.236 &1.203 & 1.197\\
		&  &0.7   &1.204 &1.197 &1.216 & 1.183\\
		&   &0.8   &1.245 & 1.221 &1.229 & 1.201\\
		&  & 0.9 &1.304 & 1.279 &1.298 & 1.274\\
		\cline{2-7}
		\hline\hline
		\multirow{8}{*}{8:2} &
		\multirow{4}{*}{0.9} &0.6   &1.108& 1.116 &1.062 & 0.995\\
		&  &0.7   &1.103 & 1.089 &1.004 & 0.982\\
		&   &0.8   &1.114& 1.120 &1.027 & 1.006\\
		&  & 0.9 &1.223& 1.187 &1.104 & 1.082\\
		\cline{2-7}
		&\multirow{4}{*}{0.8}
		&0.6   &1.482 & 1.467 &1.430 & 1.424\\
		&  &0.7   &1.469 & 1.442 &1.427 & 1.403\\
		&   &0.8   &1.487 & 1.472 & 1.448 & 1.429\\
		&  & 0.9 &1.550 &1.503 &1.523 & 1.497\\
		\cline{2-7}
		\hline
	\end{tabular}
\end{table}

\begin{table}
	\centering
	\caption{Average RMSE of $10$ trials by different sampling strategies on Jester dataset under different parameters setting.}
	\label{tab:expjester}
	\begin{tabular}{ l|l|l|lll l}
		\hline
		$\varrho_1$  &$\varrho_2$       &$\theta$        &RBF-uni           &RBF-lev   &HQASD-uni  &HQASD-lev\\
		
		\hline
		\multirow{8}{*}{Jester-1} &
		\multirow{4}{*}{0.9}
		   &0.6   &4.4921 &4.3876 &4.3805 &4.3778\\
		&  &0.7   &4.3672 &4.3600 &4.3654 &4.3580\\
		&  &0.8   &4.3801 &4.3789 &4.3799 &4.3724\\
		&  &0.9   &4.4035 &4.3908 &4.3896 &4.3861\\
		\cline{2-7}
		&\multirow{4}{*}{0.8}
		   &0.6   &5.5472 &5.4767 &5.5203 &5.4508\\
		&  &0.7   &5.4683 &5.4314 &5.4401 &5.3986\\
		&  &0.8   &5.4961 &5.4843 &5.4872 &5.4637\\
		&  &0.9   &5.5036 &5.4827 &5.4907 &5.4493\\
		\cline{2-7}
		\hline
				\hline
		\multirow{8}{*}{Jester-2} &
		\multirow{4}{*}{0.9}
		   &0.6   &4.5012 &4.4852 &4.3814 &4.3751\\
		&  &0.7   &4.3704 &4.3687 &4.3665 &4.3604\\
		&  &0.8   &4.3874 &4.3852 &4.3869 &4.3788\\
		&  &0.9   &4.4420 &4.4301 &4.3971 &4.3952\\
		\cline{2-7}
		&\multirow{4}{*}{0.8}
     	   &0.6   &5.5362 &5.4699 &5.5187 &5.4556\\
		&  &0.7   &5.4691 &5.4403 &5.4337 &5.4207\\
		&  &0.8   &5.4903 &5.4884 &5.4890 &5.4605\\
		&  &0.9   &5.5174 &5.5062 &5.5031 &5.4937\\
		\cline{2-7}
        \hline
		\hline
		\multirow{8}{*}{Jester-3} &
		\multirow{4}{*}{0.9}
		&0.6   &5.9731 &5.9720 &5.9657 &5.9563\\
		&  &0.7   &5.9080 &5.8981 &5.8673 &5.8600\\
		&  &0.8   &5.9576 &5.9097 &5.9418 &5.9385\\
		&  &0.9   &5.9604 &5.9537 &5.9554 &5.9462\\
		\cline{2-7}
		&\multirow{4}{*}{0.8}
		&0.6   &6.6831 &6.5903 &6.6531 &6.5837\\
		&  &0.7   &6.6063 &6.4772 &6.5781 &6.5174\\
		&  &0.8   &6.6605 &6.6112 &6.6417 &6.5978\\
		&  &0.9   &6.7017 &6.6508 &6.6984 &6.6003\\
		\cline{2-7}
		\hline
		\end{tabular}
\end{table}

\section{Conclusion}\label{sec:6conclusion}
The incoherence condition presents a challenge in many real-world scenarios. To address this challenge, we propose a biased sampling processing method based on the row and column leverage scores of the underlying matrix. We demonstrate that an unknown $n\times n$ matrix of rank $r$ can be exactly recovered from approximately $\mathcal{O}(nr\log^2 (n))$ entries, even in cases where some entries are corrupted. Numerical experiments support our theoretical results and demonstrate the effectiveness of the biased sampling processing.

We propose a leverage score-based biased sampling strategy for matrix completion with noise. Our analysis of the sampling upper bound is rigorous. However, the sampling lower bound remains an open problem worthy of exploration. In addition, developing other methods for estimating leverages and tuning the sampling procedure would be interesting. The extension of the results and techniques presented in this paper for matrix completion has potential implications for broader fields and is therefore of independent interest.

\ifCLASSOPTIONcaptionsoff
  \newpage
\fi

\bibliographystyle{IEEEtran}
\bibliography{IEEEexample}

%


\appendices

\section{Proof of $\mu{(\infty,2)}$ and $\mu{(\infty)}$ are norm}\label{app:normproof}
\emph{Proof.}
The $\mu{(\infty,2)}$-norm and $\mu{(\infty)}$-norm of a matrix $Z\in\mathbb{R}^{m\times n}$ is defined as
\begin{equation}
	\|Z\|_{\mu(\infty,2)}:=\max_{a,b}\{\sqrt{\frac{m}{\mu_ar}}\|Z_{a\cdot}\|_2, \sqrt{\frac{n}{\nu_br}}\|Z_{\cdot b}\|_2\}, \label{equ:muinfty2}
\end{equation}
and
\begin{equation}
	\|Z\|_{\mu{(\infty)}}:=\max_{a,b}|Z_{ab}|\sqrt{\frac{m}{\mu_a r}}\sqrt{\frac{n}{\nu_b r}}. \label{equ:muinfty}
\end{equation}
\textit{(1) Prove that $\mu{(\infty,2)}$ is a norm}\\
We first prove that $\|Z_{a\cdot}\|_2$ is a norm. To prove this, we merely check whether the three conditions of a norm (nonnegativity, homogeneity, and triangle inequality) are met. It is easy to check that $\|Z_{a\cdot}\|_2$ satisfies the nonnegative property of a norm. From \eqref{equ:muinfty2}, we know that $Z_{a\cdot}\in\mathbb{R}^n, a = 1,2,\ldots,m$. Let $\alpha\in\mathbb{R}$, then we have
\begin{align*}
	\|\alpha Z_{a\cdot}\|_2 &= \sqrt{|\alpha Z_{1\cdot}|^2+|\alpha Z_{2\cdot}|^2+\ldots+|\alpha Z_{m\cdot}|^2}\\
	& = \sqrt{|\alpha|^2|Z_{1\cdot}|^2+|\alpha|^2|Z_{2\cdot}|^2+\ldots+|\alpha|^2|Z_{m\cdot}|^2}\\
	& = \sqrt{|\alpha|^2|(Z_{1\cdot}|^2+|Z_{2\cdot}|^2+\ldots+|Z_{m\cdot}|^2)}\\
	& = |\alpha|\sqrt{|Z_{1\cdot}|^2+|Z_{2\cdot}|^2+\ldots+|Z_{m\cdot}|^2}\\
	& =  |\alpha|\|Z_{a\cdot}\|_2.
\end{align*}
This implies that $\|Z_{a\cdot}\|_2$ satisfies the homogeneity property of a norm. Let $Z_{i\cdot},Z_{j\cdot}\in\mathbb{R}^n$, then
\begin{align*}
	\|Z_{i\cdot}+Z_{j\cdot}\|_2^2 &=  (Z_{i\cdot}+Z_{j\cdot})^*(Z_{i\cdot}+Z_{j\cdot}) \\
	& = Z_{i\cdot}^*Z_{i\cdot}+Z_{i\cdot}^*Z_{j\cdot}+Z_{j\cdot}^*Z_{i\cdot}+Z_{j\cdot}^*Z_{j\cdot}\\
	& \leq \|Z_{i\cdot}\|_2^2 + \|Z_{j\cdot}\|_2^2 + 2\|Z_{i\cdot}\|_2\|Z_{j\cdot}\|_2\\
	& = \|Z_{i\cdot}\|_2^2 + \|Z_{j\cdot}\|_2^2.
\end{align*}
Taking the square root of both sides yields that $\|Z_{a\cdot}\|_2$ satisfies the triangle inequality. Thus, we get that $\|Z_{a\cdot}\|_2$ is a norm. \\
$\|Z_{b\cdot}\|_2$ is a norm can be obtained along the same lines as $\|Z_{a\cdot}\|_2$. Combining the fact that $\sqrt{\frac{m}{\mu_ar}},a=1,2,\ldots,m$ and $\sqrt{\frac{n}{\nu_br}},b=1,2,\ldots,n$ are different real numbers, we know that $\|Z\|_{\mu(\infty,2)}$ is essentially an $\ell_\infty$ norm on $\mathbb{R}^{mn}$. This completes the proof.\\
\textit{(2) Prove that $\mu{(\infty)}$ is a norm}\\
First, we can see that $\sqrt{\frac{m}{\mu_ar}},a=1,2,\ldots,m$ and $\sqrt{\frac{n}{\nu_br}},b=1,2,\ldots,n$ are different real numbers. From the definition of $\mu{(\infty)}$ in \eqref{equ:muinfty}, we know that $\|Z\|_{\mu(\infty,2)}$ is essentially an $\ell_\infty$ norm on $\mathbb{R}^{mn}$.

\section{Proof of Theorem \ref{thm:leverageclose}}\label{app:a}
\emph{Proof.} 
$\delta< 1$ ensures that leverage scores of $X$ are well defined. Let $\varepsilon = \frac{\|S\|}{\|L\|}$ and $\varepsilon^\bot = \frac{\|(I-LL^\dag)S\|}{\|L\|}$.
Then based on the principal angle theory (Sec. 2.1 in \cite{Holodnak20151143}), we know that $\varepsilon^\bot$ removes the contribution of $S$ to some extent that lies in $\text{Range}(L)$. When $\varepsilon$ is large, $S$ has only a small contribution in $\text{Range}(L)$. Note that $S$ does not change the leverage scores if $\text{Range}(L) = \text{Range}(X)$. Thus, for a given matrix, its leverage scores calculated by SVD and QR hold the same characteristics. Then by Theorem 2.4 in \cite{Holodnak20151143}, (\ref{equ:nearmu}) is obtained. Considering $X^*$ and $L^*$, we can also obtain the (\ref{equ:nearnu}) in the same manner.

\section{Proof of Lemma \ref{lem:duallem}}\label{app:b}
\emph{Proof.} Set $\hat{L}=L+H$. Due to the fact that $S$ is supported by $\Omega$ and $\mathcal{P}_O(L)+S=\mathcal{P}_O(\hat L)+\hat S$, we have $\mathcal{P}_O(\hat L-L)=\mathcal{P}_O(H)=\mathcal{P}_O(S-\hat S)$.

By the subgradient of the unclear norm at $L$, we have
\begin{align}
	\|L+H\|_*&\geq\|L\|_*+\langle UV^*+\Delta_1,H\rangle \nonumber\\
	&=\|L\|_*+\langle UV^*,H\rangle + \langle\Delta_1,\mathcal P_{T^\bot}(H)\rangle\nonumber\\
	&\overset{(a)}{=}\|L\|_*+\langle UV^*,H\rangle + \|\mathcal P_{T^\bot}(H)\|_*,\label{equ:i1}
\end{align}
where (a) follows the fact that there exists a $\Delta_1$ and $\|\mathcal P_{T^\bot}\Delta_1\|\leq 1$ such that $\|\mathcal P_{T^\bot}(H)\|_* =\langle\Delta_1,\mathcal P_{T^\bot}(H)\rangle $.

Because
$(\hat{L},\hat{S})$ is the optimal solution,
\begin{equation}
	\|L\|_*+\lambda\|S\|_1 \geq \|\hat L\|_*+\lambda\|\hat S\|_1.\label{equ:e1}
\end{equation}
From (\ref{equ:i1}) and (\ref{equ:e1}), we obtain
\begin{equation*}
	\lambda\|S\|_1-\lambda\|\hat{S}\|_1\geq\langle UV^*, H\rangle + \|\mathcal P_{T^\bot} H\|_*.
\end{equation*}
This implies
\begin{align*}
	\lambda\|S\|_1-\lambda\|\mathcal P_{O/\Gamma^{'}}\hat S\|_1&\geq\langle UV^*, H\rangle + \|\mathcal P_{T^\bot} H\|_* \\&+\lambda\|\mathcal P_{\Gamma^{'}}\hat S\|_1.
\end{align*}
On the other hand,
\begin{align*}
	\|\mathcal P_{O/\Gamma^{'}}\hat S\|_1&=\|S-\mathcal P_{O/\Gamma^{'}}H\|\\
	&\geq\|S\|_1 + \langle\text{sgn}(S)+\Delta_2,\mathcal P_{O/\Gamma^{'}}(-H)\rangle\\
	&=\|S\|_1 + \langle\text{sgn}(S)+\Delta_2,\mathcal P_{O/(\Gamma^{'}\cup\Omega)}(-H)\\&+\mathcal P_{\Omega}(-H)\rangle\\
	&\overset{(a)}{=}\|S\|_1 + \langle\text{sgn}(S),\mathcal P_{\Omega}(-H)\rangle\\&+\|\mathcal P_{O/(\Gamma^{'}\cup\Omega)}(-H)\|_1\\
	&\geq \|S\|_1 + \langle-H, \mathcal{P}_{O/\Gamma^{'}}(W)\rangle,
\end{align*}

where (a) follows the fact that there exists a $\Delta_2$ and $\|\mathcal{P}_{\Omega^c}(\Delta_2)\|_\infty\leq1$ such that $\langle\Delta_2,\mathcal P_{O/(\Gamma^{'}\cup\Omega)}(-H)\rangle=\|\mathcal P_{O/(\Gamma^{'}\cup\Omega)}(-H)\|_1$.

By the above two inequalities, it yields
\begin{equation*}
	\langle H,\lambda\mathcal{P}_{O/\Gamma^{'}}(W)-UV^*\rangle\geq\|\mathcal{P}_{T^\bot}(H)\|_*+\lambda\|\mathcal{P}_{\Gamma^{'}}(\hat{S})\|_1.
\end{equation*}
Besides,
\begin{align*}\label{equ:i3}
	&\langle H,\lambda\mathcal{P}_{O/\Gamma^{'}}(W)-UV^*\rangle \\
	&= \langle H,\lambda\mathcal{P}_{O/\Gamma^{'}}(W)-UV^*+Y\rangle - \langle H,Y\rangle\\
	&=\langle \mathcal{P}_T(H),\mathcal{P}_T(\lambda\mathcal{P}_{O/\Gamma^{'}}(W)-UV^*+Y)\rangle\\&+\langle \mathcal{P}_{T^\bot}(H),\mathcal{P}_{T^\bot}(\lambda\mathcal{P}_{O/\Gamma^{'}}(W)-Y)\rangle\\
	&-\langle\mathcal{P}_{\Gamma^{'}}(H),\mathcal{P}_{\Gamma^{'}}(Y)\rangle-\langle\mathcal{P}_{\Gamma^{'^c}}(H),\mathcal{P}_{\Gamma^{'^c}}(Y)\rangle\\
	&{\leq} \frac{\lambda}{n^3}\|\mathcal{P}_T(H)\|_F +\frac{1}{4}\|\mathcal{P}_{T^\bot}(H)\|_*+\frac{\lambda}{4}\|\mathcal{P}_{\Gamma^{'}}(H)\|_1.
\end{align*}

Then we have
\begin{equation}\label{equ:i4}
	\frac{3}{4}\|\mathcal{P}_{T^\bot}(H)\|_*+\frac{3\lambda}{4}\|\mathcal{P}_{\Gamma^{'}}(H)\|_1\geq\frac{\lambda}{n^3}\|\mathcal{P}_T(H)\|_F.
\end{equation}

By Lemma 3, we have $\|\frac{1}{(1-2q)p_{ij}}\mathcal{P}_T\mathcal P_{\Gamma^{'}}\mathcal{P}_T-\mathcal{P}_T\|\leq\frac{1}{2}$ and $\|\frac{1}{\sqrt{(1-2q)p_{ij}}}\mathcal{P}_T\mathcal P_{\Gamma^{'}}\|\leq\sqrt\frac{3}{2}$.

Then,
\begin{align*}
	\|\mathcal{P}_T(H)\|_F&\leq 2\|\frac{1}{(1-2q)p_{ij}}\mathcal{P}_T  \mathcal P_{\Gamma^{'}}  \mathcal{P}_T(H)\|_F \\
	&\leq 2\|\frac{1}{(1-2q)p_{ij}}\mathcal{P}_T  \mathcal P_{\Gamma^{'}}  \mathcal{P}_{T^\bot}(H)\|_F \\&+ 2\|\frac{1}{(1-2q)p_{ij}}\mathcal{P}_T  \mathcal P_{\Gamma^{'}}(H)\|_F\\
	&\leq~ \sqrt{\frac{6}{(1-2q)p_{ij}}}\|\mathcal{P}_{T^\bot}(H)\|_F\\&+\sqrt{\frac{6}{(1-2q)p_{ij}}}\|\mathcal{P}_{\Gamma^{'}}(H)\|_F.
\end{align*}
From the above inequality and (\ref{equ:i4}), we obtain
\begin{align}
	&(\frac{3}{4}-\frac{\lambda}{n^3}\sqrt{\frac{6}{(1-2q)p_{ij}}})\|\mathcal{P}_{T^\bot}(H)\|_F\\&+(\frac{3\lambda}{4}-\frac{\lambda}{n^3}\sqrt{\frac{6}{(1-2q)p_{ij}}})\|\mathcal{P}_{\Gamma^{'}}(H)\|_F\leq 0.
\end{align}
The above inequality always holds if $p_{ij}\geq\frac{1}{n^5}$. This implies $\|\mathcal{P}_{T^\bot}(H)\|_F=\|\mathcal{P}_{\Gamma^{'}}(H)\|_F$, which further implies $\|\mathcal{P}_{\Gamma^{'}}\mathcal{P}_T(H)\|_F=0$. Since $\|\frac{1}{(1-2q)p_{ij}}\mathcal{P}_T\mathcal P_{\Gamma^{'}}\mathcal{P}_T-\mathcal{P}_T\|\leq\frac{1}{2}$, we know that $\mathcal{P}_{\Gamma^{'}}\mathcal{P}_T$ is injective on $T$. We then have $\mathcal{P}_T(H)=0$. Hence, $H=0$. This completes the proof.

\section{Proof of Lemma \ref{lem:lem3}}\label{app:c}
\emph{Proof.} Under the assumption that $Z\in\text{Range}(\mathcal{P}_T)$, we have
\begin{align*}
	&\|(\frac{1}{(1-2q)p_{ij}}\mathcal{P}_T\mathcal P_{\Gamma}\mathcal{P}_T-\mathcal{P}_T)Z\|_{\mu(\infty,2)}\\
	&=\|(\frac{1}{(1-2q)p_{ij}}\mathcal{P}_T\mathcal P_{\Gamma}-\mathcal{P}_T)Z\|_{\mu(\infty,2)}.
\end{align*}
Then by Lemma 11 in \cite{Chen20152999}, we obtain the desired results.

\section{Proof of Lemma \ref{lem:lem4}} \label{app:d}
\emph{Proof.} Under the assumption that $Z\in\text{Range}(\mathcal{P}_T)$, we have
\begin{align*}
	&\|(\frac{1}{(1-2q)p_{ij}}\mathcal{P}_T\mathcal P_{\Gamma}\mathcal{P}_T-\mathcal{P}_T)Z\|_{\mu(\infty)}\\
	&=\|(\frac{1}{(1-2q)p_{ij}}\mathcal{P}_T\mathcal P_{\Gamma}-\mathcal{P}_T)Z\|_{\mu(\infty)}.
\end{align*}
Then by Lemma 12 in \cite{Chen20152999}, we obtain the desired results.

\section{Validity of the Dual Certificate} \label{app:e}
\noindent \textbf{Validating inequality (\ref{equ:cond1})}.
We first bound each elements of $\mathcal{P}_T\mathcal{P}_{\Omega^{'}}(W)$ and the Frobenius norm of $X_0$.

For any given index pair $(a,b)\in[n]\times[n]$,
\begin{align*}
	[\mathcal{P}_T\mathcal{P}_{\Omega^{'}}(W)]_{ab}
	&=\langle e_ae_b^*,\mathcal{P}_T\mathcal{P}_{\Omega^{'}}(W)\rangle\\
	&=\langle \mathcal{P}_T(e_ae_b^*),\mathcal{P}_{\Omega^{'}}(W)\rangle\\
	&=\sum_{i,j}\delta_{ij}\langle e_ie_j^*,\mathcal{P}_T(e_ae_b^*)\rangle\\
	&:=\sum_{i,j}S_{ij},
\end{align*}
where
\begin{align}\label{equ:delta}
	\delta_{ij}=\left\{
	\begin{array}{lll}
		\frac{p_{ij}q}{1-p_{ij}+2qp_{ij}},\quad if ~W_{ij}=1,\\
		\frac{1-p_{ij}}{1-p_{ij}+2qp_{ij}},\quad if ~W_{ij}=0,\\
		\frac{p_{ij}q}{1-p_{ij}+2qp_{ij}},\quad if ~W_{ij}=-1.
	\end{array}
	\right.
\end{align}

Clearly, $\mathbb{E}(S_{ij})=0$ and for all $(i,j)\in[n]\times[n]$, $S_{ij}$ are independent random variables. Note that
\begin{align*}
	|S_{ij}|&\leq|\langle e_ie_j^*,\mathcal{P}_T(e_ae_b^*)\rangle|\\
	&=|e_i^*UU^*e_ae_b^*e_j + e_i^*e_ae_b^*VV^*e_j \\
	&- e_i^*UU^*e_ae_b^*VV^*e_j|.
\end{align*}
If $i=a,j=b$, we have
\begin{align*}
	&|\langle e_ie_j^*,\mathcal{P}_T(e_ae_b^*)\rangle|\\
	&=|e_a^*UU^*e_a + e_a^*(I-UU^*)e_ae_b^*VV^*e_b|\\
	&\leq\|e_a^*U\|_F^2+\|e_a^*V\|_F^2\leq\frac{(\mu_a+\nu_b)r}{n}.
\end{align*}

If $i=a,j\neq b$, we have
\begin{align*}
	&|\langle e_ie_j^*,\mathcal{P}_T(e_ae_b^*)\rangle|\\
	&=|e_a^*(I-UU^*)e_ae_b^*VV^*e_j|\\
	&\leq |e_b^*VV^*e_j|\leq\sqrt{\frac{\nu_br}{n}}\sqrt{\frac{\nu_jr}{n}}.
\end{align*}

If $i\neq a,j= b$, we have
\begin{align*}
	&|\langle e_ie_j^*,\mathcal{P}_T(e_ae_b^*)\rangle|\\
	&=|e_i^*UU^*e_ae_b^*(I-VV^*)e_b|\\
	&\leq|e_i^*UU^*e_a|\leq\sqrt{\frac{\mu_ar}{n}}\sqrt{\frac{\mu_ir}{n}}.
\end{align*}

If $i\neq a,j\neq b$, we have
\begin{align*}
	&|\langle e_ie_j^*,\mathcal{P}_T(e_ae_b^*)\rangle|=|e_i^*UU^*e_ae_b^*VV^*e_j|\\
	&\leq|e_i^*UU^*e_a||e_b^*VV^*e_j|\\
	&\leq\sqrt{\frac{\mu_ar}{n}}\sqrt{\frac{\mu_ir}{n}}\sqrt{\frac{\nu_br}{n}}\sqrt{\frac{\nu_jr}{n}}.
\end{align*}
Thus, we conclude that
\begin{equation*}
	|S_{ij}|\leq\sqrt{\frac{2\mu r}{n}}\sqrt{\frac{(\mu_a+\nu_b)r}{2}},
\end{equation*}
where $\mu = \max\{\mu_i,\nu_j\}$.

On the other hand, note that
\begin{align*}
	|\sum_{i,j}\mathbb{E}(S_{ij}^2)|
	&=|\sum_{i,j}\mathbb{E}\delta_{ij}\langle e_ie_j^*,\mathcal{P}_T(e_ae_b^*)\rangle^2|\\
	&=|\sum_{i,j}\frac{2p_{ij}q}{1-p_{ij}+2qp_{ij}}\langle e_ie_j^*,\mathcal{P}_T(e_ae_b^*)\rangle^2|\\
	&\leq|\sum_{i,j}\langle e_ie_j^*,\mathcal{P}_T(e_ae_b^*)\rangle^2|\\
	&=\|\mathcal P_{T}(e_ae_b^*)\|_F^2\\
	&\leq\frac{(\mu_a+\nu_b)r}{n}.
\end{align*}
By Bernstein Inequality in Lemma \ref{lem:bernstein}, we obtain
\begin{align}\label{equ:i6}
	[\mathcal{P}_T\mathcal{P}_{\Omega^{'}}(W)]_{ab}&=|\sum_{i,j}S_{ij}| \nonumber\\
	&\leq C\sqrt{\frac{(\mu_a+\nu_b)r}{n}\log n}.
\end{align}
We now turn to bound $\|X_0\|_F$.
\begin{align*}
	\|X_0\|_F&=\|\mathcal{P}_T(UV^*-\lambda\mathcal{P}_{\Omega^{'}}(W))\|_F\\
	&\leq \|UV^*\|_F+\lambda\|\mathcal{P}_{\Omega^{'}}(W))\|_F\\
	&\leq \sqrt{r}+C'\max_{a,b}\sqrt{(\mu_a+\nu_b)r}\\
	&= C\max_{a,b}\sqrt{(\mu_a+\nu_b)r}.
\end{align*}
We next validate inequality (\ref{equ:cond1}).
\begin{align*}
	&\|\mathcal{P}_T(Y)+\mathcal{P}_T(\lambda\mathcal{P}_{\Omega^{'}}(W)-UV^*)\|_F\\
	&=\|-X_0+\sum_{k=1}^{t}\frac{1}{\rho_k}(\mathcal P_T\mathcal P_{\Gamma_{k}})X_{k-1}\|_F\\
	&\overset{(a)}{=}\|\mathcal P_T(X_0)-\sum_{k=1}^{t}\frac{1}{\rho_k}(\mathcal P_T\mathcal P_{\Gamma_{k}}\mathcal P_T)X_{k-1}\|_F
\end{align*}
\begin{align*}
	&=\|\mathcal (P_T-\frac{1}{\rho_1}\mathcal P_T\mathcal P_{\Gamma_{1}}\mathcal P_T)X_0\\
	&-\sum_{k=2}^{t}\frac{1}{\rho_k}(\mathcal P_T\mathcal P_{\Gamma_{k}}\mathcal P_T)X_{k-1}\|_F\\
	&=\|X_1-\sum_{k=2}^{t}\frac{1}{\rho_k}(\mathcal P_T\mathcal P_{\Gamma_{k}}\mathcal P_T)X_{k-1}\|_F\\
	&=\ldots=\|X_t\|_F\overset{(b)}{\leq}(\frac{1}{2})^t\|X_0\|_F\\
	&\leq C(\frac{1}{2})^t\max_{a,b}\sqrt{(\mu_a+\nu_b)r}\leq \frac{\lambda}{n^3},
\end{align*}
where (a) follows from (\ref{equ:e2}), (b) follows from (\ref{equ:i5}).

\noindent\textbf{Validating inequality (\ref{equ:cond2})}.
We first bound $\|X_0\|_{\mu(\infty)}$ and $\|X_0\|_{\mu(\infty,2)}$.
\begin{align}
	\|X_0\|_{\mu(\infty)}&=\|\mathcal{P}_T(UV^*-\lambda\mathcal{P}_{\Omega^{'}}(W))\|_{\mu(\infty)}\nonumber\\
	&=\|UV^*-\lambda\mathcal{P}_T\mathcal{P}_{\Omega^{'}}(W)\|_{\mu(\infty)}\nonumber\\
	&\leq \|UV^*\|_{\mu(\infty)}+\lambda\|\mathcal{P}_T\mathcal{P}_{\Omega^{'}}(W)\|_{\mu(\infty)}\nonumber\\
	&\overset{(a)}{\leq}1+\lambda\max_{a,b}\sqrt{\frac{(\mu_a+\nu_b)r}{n}\log n}\sqrt{\frac{n^2}{\mu_a\nu_br^2}}\nonumber\\
	&\leq\frac{c}{\sqrt{r}}+1\leq c',\label{equ:i7}
\end{align}
where (a) follows from (\ref{equ:i6}) and $\|UV^*\|_{\mu(\infty)}\leq 1$.
\begin{align*}
	\|X_0\|_{\mu(\infty,2)}&=\|\mathcal{P}_T(UV^*-\lambda\mathcal{P}_{\Omega^{'}}(W))\|_{\mu(\infty,2)}\\
	&\leq\|UV^*\|_{\mu(\infty,2)}+\lambda\|\mathcal{P}_T\mathcal{P}_{\Omega^{'}}(W)\|_{\mu(\infty,2)}.
\end{align*}
It is easy to verify that $\|UV^*\|_{\mu(\infty,2)}=1$. We now focus on bounding $\|\mathcal{P}_T\mathcal{P}_{\Omega^{'}}(W)\|_{\mu(\infty,2)}$.

Let $B=\mathcal{P}_T\mathcal{P}_{\Omega^{'}}(W)$, by the definition of the $\mu(\infty,2)$-norm, we bound each term. Note that
\begin{align*}
	\sqrt{\frac{n}{\nu_br}}B_{\cdot b}
	&=\sum_{i,j}(B_{ij}\mathcal{P}_T(e_ie_j^*)e_b)\delta_{ij}\sqrt{\frac{n}{\nu_br}}\\
	&:=\sum_{i,j}S_{ij},
\end{align*}
where $\delta_{ij}$ is defined in (\ref{equ:delta}) and $\mathbb{E}(S_{ij})=0$. This implies that $\sqrt{\frac{n}{\nu_br}}B_{\cdot b}$ can be written as the sum of independent column vectors. To use Bernstein Inequality, we should control $|S_{ij}|$ and $|\sum_{i,j}\mathbb{E}(S_{ij}^2)|$. We first bound $\|\mathcal{P}_T(e_ie_j^*)e_b\|$.

If $j=b$, we have
\begin{align*}
	&\|\mathcal{P}_T(e_ie_j^*)e_b\|\\
	&=\|UU^*e_i+e_ie_j^*VV^*e_b-UU^*e_ie_j^*VV^*e_b\|\\
	&\leq\|UU^*e_i\|+\|(I-UU^*)e_ie_j^*VV^*e_b\|\\
	&\leq\|UU^*e_i\|+\|V^*e_b\|^2\\
	&\leq\sqrt{\frac{\mu_ir}{n}}+\sqrt{\frac{\nu_br}{n}}\leq\sqrt{\frac{2(\mu_i+\nu_b)r}{n}}.
\end{align*}
If $j\neq b$, we have
\begin{align*}
	\|\mathcal{P}_T(e_ie_j^*)e_b\|&=\|(I-UU^*)e_ie_j^*VV^*e_b\|\\
	&\leq\|e_j^*VV^*e_b\|\leq\|VV^*e_b\|\\
	&\leq\sqrt{\frac{\nu_br}{n}}.
\end{align*}
Thus we obtain that for $j=b$,
\begin{align*}
	|S_{ij}|&\leq|B_{ij}|\sqrt{\frac{2(\mu_i+\nu_b)r}{n}}\sqrt{\frac{n}{\nu_br}}\\
	&\leq C|B_{ij}|.
\end{align*}
For $j\neq b$,
\begin{align*}
	|S_{ij}|&\leq|B_{ij}|\sqrt{\frac{\nu_br}{n}}\sqrt{\frac{n}{\nu_br}}=|B_{ij}|.
\end{align*}
Therefore, by (\ref{equ:i6}), we conclude that $|S_{ij}|\leq|B_{ij}|\leq C\max_{i,j}\sqrt{\frac{(\mu_i+\nu_j)r}{n}\log n}$.

On the other hand,
\begin{align*}
	|\sum_{i,j}\mathbb{E}(S_{ij}^2)|&=\sum_{i,j}\mathbb{E}\delta_{ij}^2|B_{ij}|^2\|\mathcal{P}_T(e_ie_j^*)e_b\|^2\frac{n}{\nu_br}\\
	&=\sum_{i,j}\frac{2p_{ij}q|B_{ij}|^2}{1-p_{ij}+2qp_{ij}}\|\mathcal{P}_T(e_ie_j^*)e_b\|^2\frac{n}{\nu_br}\\
	&\leq\sum_{i,j}|B_{ij}|^2\|\mathcal{P}_T(e_ie_j^*)e_b\|^2\frac{n}{\nu_br}.
\end{align*}
Thus, we obtain that for $j=b$,
\begin{align*}
	&|\sum_{i,j}\mathbb{E}(S_{ij}^2)|\leq|B_{ij}|^2\frac{n}{\nu_br}{\frac{2(\mu_i+\nu_b)r}{n}}\\&\leq C|B_{ij}|^2.
\end{align*}
For $j\neq b$,
\begin{align*}
	|\sum_{i,j}\mathbb{E}(S_{ij}^2)|\leq|B_{ij}|^2\frac{n}{\nu_br}{\frac{\nu_br}{n}}=|B_{ij}|^2.
\end{align*}
Therefore, we conclude that $|\sum_{i,j}\mathbb{E}(S_{ij}^2)|\leq|B_{ij}|^2\leq C\max_{i,j}\frac{(\mu_i+\nu_j)r}{n}\log n$.

Applying Matrix Bernstein Inequality in Lemma \ref{lem:bernstein}, with high probability, we have
\begin{align*}
	\sqrt{\frac{n}{\nu_br}}\|B_{\cdot b}\|=\|\sum_{i,j}S_{ij}\|&\leq C|B_{ij}|(\sqrt{\log n}+\log n)\\
	&\leq C'\log n.
\end{align*}

We now proceed to bound $\|X_0\|_{\mu(\infty,2)}$
\begin{align}
	\|X_0\|_{\mu(\infty,2)}&\leq\|UV^*\|_{\mu(\infty,2)}+\lambda\|\mathcal{P}_T\mathcal{P}_{\Omega^{'}}(W)\|_{\mu(\infty,2)},\nonumber\\
	&\leq 1+C'\lambda\log n\leq C.\label{equ:i8}
\end{align}
In order to show that $Y$ satisfies (\ref{equ:cond2}), we bound $\|\mathcal{P}_{T^\bot}(Y)\|$ and $\|\mathcal{P}_{T^\bot}(\lambda\mathcal{P}_{\Omega^{'}}(W))\|$ separately.
\begin{align*}
	\|\mathcal{P}_{T^\bot}(\lambda\mathcal{P}_{\Omega^{'}}(W))\|\leq \lambda\|\mathcal{P}_{\Omega^{'}}(W)\|\overset{(a)}{\leq}C\lambda \sqrt{n}\leq \frac{1}{8},
\end{align*}
where (a) follows from the spectral norm bound on random matrix in \cite{Eldar2012}.
\begin{align*}
	\|\mathcal{P}_{T^\bot}(Y)\|&=\|\mathcal{P}_{T^\bot}\sum_{k=1}^{t}\frac{1}{\rho_k}\mathcal P_{\Gamma_{k}}X_{k-1}\|\\
	&\leq\sum_{k=1}^{t}\|\frac{1}{\rho_k}\mathcal{P}_{T^\bot}\mathcal P_{\Gamma_{k}}X_{k-1}\|\\
	&\overset{(a)}{=}\sum_{k=1}^{t}\|\mathcal{P}_{T^\bot}(\frac{1}{\rho_k}\mathcal P_{\Gamma_{k}}X_{k-1}-X_{k-1})\|\\
	&\leq\sum_{k=1}^{t}\|\frac{1}{\rho_k}\mathcal P_{\Gamma_{k}}X_{k-1}-X_{k-1}\|\\
	&\overset{(b)}{\leq}\frac{c}{\sqrt{c_0}}\sum_{k=1}^{t}(\|X_{k-1}\|_{\mu(\infty,2)}+\|X_{k-1}\|_{\mu(\infty)}),
\end{align*}
where $c>1$ and $c_0$ is sufficiently large; (a) follows from (\ref{equ:e2}); (b) follows Lemma 10 in \cite{Chen20152999}.

By Lemma \ref{lem:lem4}, we obtain
\begin{align*}
	\|X_{k-1}\|_{\mu{(\infty)}}\leq(\frac{1}{2})^{k-1}\|X_{0}\|_{\mu{(\infty)}}.
\end{align*}
By Lemma \ref{lem:lem3}, we obtain

\begin{align*}
	\|X_{k-1}\|_{\mu(\infty,2)}&\leq\frac{1}{2}\|X_{k-2}\|_{\mu{(\infty)}}+\frac{1}{2}\|X_{k-2}\|_{\mu(\infty,2)}\\
	&\leq(\frac{1}{2})^{k-1}\|X_{0}\|_{\mu{(\infty)}}+(\frac{1}{2})^2\|X_{k-3}\|_{\mu{(\infty)}}\\
	&+(\frac{1}{2})^2\|X_{k-3}\|_{\mu(\infty,2)}\\
	&\leq 2(\frac{1}{2})^{k-1}\|X_{0}\|_{\mu{(\infty)}}+(\frac{1}{2})^3\|X_{k-4}\|_{\mu{(\infty)}}\\
	&+(\frac{1}{2})^3\|X_{k-4}\|_{\mu(\infty,2)}\\
	&\leq 3(\frac{1}{2})^{k-1}\|X_{0}\|_{\mu{(\infty)}}+(\frac{1}{2})^4\|X_{k-5}\|_{\mu{(\infty)}}\\
	&+(\frac{1}{2})^4\|X_{k-5}\|_{\mu(\infty,2)}\\
	&\leq\ldots\\
	&\leq(k-1)(\frac{1}{2})^{k-1}\|X_{0}\|_{\mu{(\infty)}}\\
	&+(\frac{1}{2})^{k-1}\|X_{0}\|_{\mu{(\infty,2)}}.
\end{align*}

Thus,
\begin{align}
	\|\mathcal{P}_{T^\bot}(Y)\|&\leq\frac{c}{\sqrt{c_0}}\sum_{k=1}^{t}k(\frac{1}{2})^{k-1}\|X_{0}\|_{\mu{(\infty)}} \nonumber\\
	&+\frac{c}{\sqrt{c_0}}\sum_{k=1}^{t}(\frac{1}{2})^{k-1}\|X_{0}\|_{\mu{(\infty,2)}}  \nonumber\\
	&\leq\frac{4c}{\sqrt{c_0}}\|X_0\|_{\mu(\infty)}+\frac{2c}{\sqrt{c_0}}\|X_0\|_{\mu(\infty,2)}  \nonumber\\
	&\overset{(a)}{\leq}\frac{1}{8},
\end{align}
where (a) follows that $c_0$ is sufficiently large and $\|X_0\|_{\mu(\infty,2)}$,$\|X_0\|_{\mu(\infty,2)}$ are both bounded (expressed in (\ref{equ:i7}) and (\ref{equ:i8})).

\noindent\textbf{Validating inequality (\ref{equ:cond3})}.\\
By Corollary \ref{cor:cor1}, we obtain that
\begin{align}\label{equ:i9}
	\|X_1\|_{\mu(\infty)}\leq\frac{1}{2\sqrt{\log n}}\|X_0\|_{\mu(\infty)},
\end{align}
and
\begin{align}\label{equ:i10}
	\|X_k\|_{\mu(\infty)}\leq\frac{1}{2^k\log n}\|X_0\|_{\mu(\infty)},\quad k=2,\ldots,t.
\end{align}
Then,
\begin{align*}
	&\|\mathcal{P}_{\Gamma^{'}}(Y)\|_\infty\\
	&\leq\|\mathcal{P}_{\Gamma^{'}}\sum_{k=1}^{t}\frac{1}{\rho_k}\mathcal P_{\Gamma_{k}}X_{k-1}\|_\infty\\
	&\leq\|\sum_{k=1}^{t}\frac{1}{\rho_k}\mathcal P_{\Gamma_{k}}X_{k-1}\|_\infty\\
	&\leq\sum_{k=1}^{t}\|\sum_{ij}\frac{1}{\rho_k}\mathbb{I}_{\{(i,j)\in\Gamma_{k}\}}(X_{k-1})_{ij}e_ie_j^*\|_\infty\\
	&\leq\max_{ij}\frac{3}{2}\frac{\|(X_{0})_{ij}\|}{\rho_1}+\sum_{k=3}^{t}\max_{ij}\frac{\|(X_{k-1})_{ij}\|}{\rho_k}\\
	&\overset{(a)}{\leq}[\frac{3}{2}\frac{1}{c_0\sqrt{n}\log n}+\sum_{k=3}^{t}\frac{1}{c_0\sqrt{n}\log n}(\frac{1}{2})^{k-1}]\|X_0\|_{\mu(\infty)}\\
	&\leq\frac{c}{c_0\sqrt{n}\log n}\leq\frac{\lambda}{4},
\end{align*}
provided $c_0$ is sufficiently large; (a) follows from (\ref{equ:i9}) and (\ref{equ:i10}); $\mathbb{I}_{\{ \cdot\}}$ is the indicator function.

\end{document}